\documentclass{article}

\PassOptionsToPackage{numbers,sort}{natbib}

\usepackage[final]{nips_2018}

\usepackage[utf8]{inputenc}    
\usepackage[T1]{fontenc}       
\usepackage{hyperref}          
\usepackage{url}               
\usepackage{amsmath}
\usepackage{amsfonts}          
\usepackage{nicefrac}          
\usepackage{microtype}         
\usepackage{bibentry}          
\usepackage{graphicx}
\usepackage{xcolor}
\usepackage{subcaption}
\usepackage{ragged2e}

\usepackage{wrapfig}

\graphicspath{{figures/}}

\nobibliography*
\DeclareMathOperator{\E}{\mathbb{E}}
\DeclareMathOperator{\Var}{Var}
\DeclareMathOperator{\Bias}{Bias}
\DeclareMathOperator{\TD}{\textrm{TD}}
\DeclareMathOperator{\MVE}{\textrm{MVE}}
\DeclareMathOperator{\STEVE}{\textrm{STEVE}}

\graphicspath{{figures/}}

\title{Sample-Efficient Reinforcement Learning\\with Stochastic Ensemble Value Expansion}

\author{
Jacob Buckman\thanks{This work was completed as part of the Google AI Residency program.} \quad Danijar Hafner \quad George Tucker \quad Eugene Brevdo \quad Honglak Lee \\
  Google Brain, Mountain View, CA, USA \\
  \texttt{jacobbuckman@gmail.com}, \;
  \texttt{mail@danijar.com},\\ \texttt{\{gjt,ebrevdo,honglak\}@google.com} \\
}

\begin{document}

\maketitle

\begin{abstract}

Integrating model-free and model-based approaches in reinforcement learning has the potential to achieve the high performance of model-free algorithms with low sample complexity. 
However, this is difficult because an imperfect dynamics model can degrade the performance of the learning algorithm, and in sufficiently complex environments, the dynamics model will almost always be imperfect. 
As a result, a key challenge is to combine model-based approaches with model-free learning in such a way that errors in the model do not degrade performance. We propose stochastic ensemble value expansion (STEVE), a novel model-based technique that addresses this issue. By dynamically interpolating between model rollouts of various horizon lengths for each individual example, STEVE ensures that the model is only utilized when doing so does not introduce significant errors. Our approach outperforms model-free baselines on challenging continuous control benchmarks with an order-of-magnitude increase in sample efficiency, and in contrast to previous model-based approaches, performance does not degrade in complex environments.
\end{abstract}

\section{Introduction}
\label{introduction}

Deep model-free reinforcement learning has had great successes in recent years, notably in playing video games~\cite{mnih2013playing} and strategic board games~\cite{silver2016mastering}. 
However, training agents using these algorithms
requires tens to hundreds of millions of samples, which makes many practical applications infeasible,
particularly in real-world control problems (e.g., robotics) where data collection is expensive.

Model-based approaches aim to reduce the number of samples required to learn a policy
by modeling the dynamics of the environment. A dynamics model can be used to
increase sample efficiency in various ways, including 
training the policy on rollouts from the dynamics model~\cite{sutton1990integrated}, 
using rollouts to improve targets for temporal difference (TD) learning~\cite{feinberg2018model},
and using information gained from rollouts as inputs to the policy~\cite{weber2017imagination}.
Model-based algorithms such as PILCO~\cite{deisenroth2011pilco} have shown that it is possible
to learn from orders-of-magnitude fewer samples.

These successes have mostly been limited to environments where the dynamics are simple
to model. In noisy, complex environments, it is difficult to learn an accurate
model of the environment. When the model makes mistakes in this context, it can cause
the wrong policy to be learned, hindering performance. Recent work has begun to address this issue.
\citet{kalweit2017uncertainty} train a model-free algorithm on a mix of real and imagined data,
adjusting the proportion in favor of real data as the Q-function becomes more confident.
\citet{kurutach2018modelensemble} train a model-free algorithm on purely imaginary data, but
use an ensemble of environment models to avoid overfitting to errors made by any individual model.

\begin{figure}[t]
\centering
\begin{subfigure}[t]{.49\textwidth}
\centering
\includegraphics[width=\textwidth]{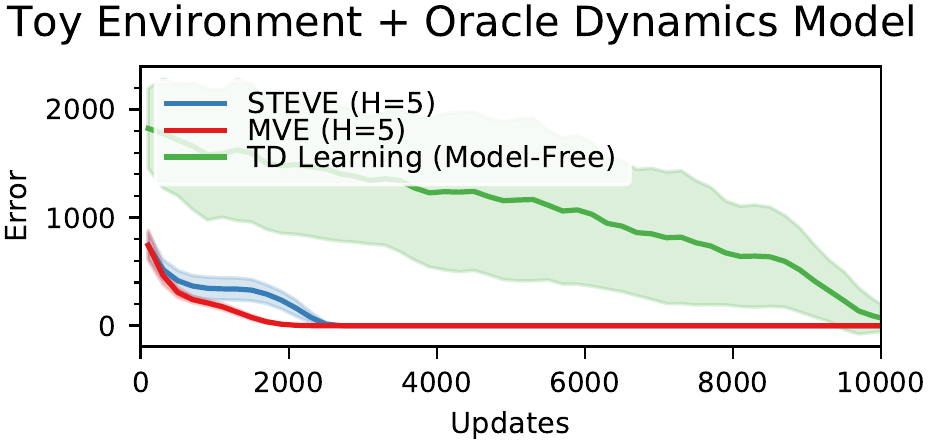}
\end{subfigure}\hfill%
\begin{subfigure}[t]{.49\textwidth}
\centering
\includegraphics[width=\textwidth]{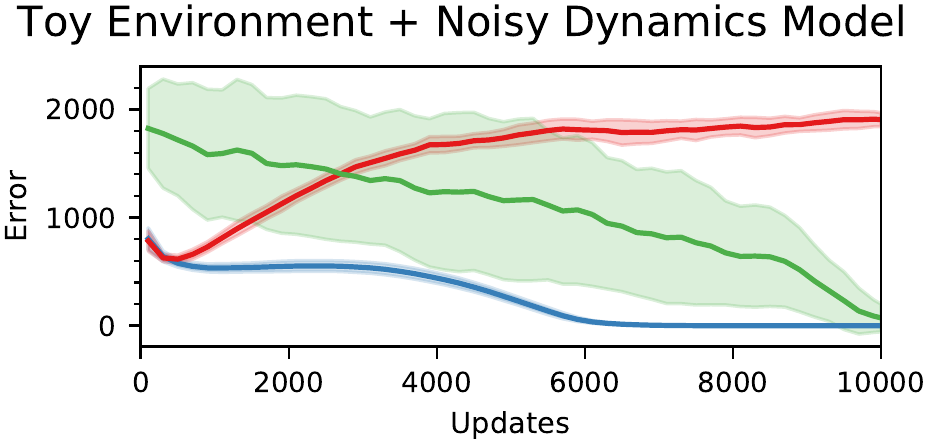}
\end{subfigure}\hfill%
\caption{Value error per update on a value-estimation task (fixed policy) in a toy environment. $H$ is the maximum rollout horizon (see Section \ref{method}). When given access to a perfect dynamics model, hybrid model-free model-based approaches (MVE and STEVE) solve this task with $5\times$ fewer samples than model-free TD learning. However, when only given access to a noisy dynamics model, MVE diverges due to model errors. In contrast, STEVE converges to the correct solution, and does so with a $2\times$ speedup over TD learning. This is because STEVE dynamically adapts its rollout horizon to accommodate model error. See Appendix \ref{toy} for more details.
}
\label{fig:toy}
\end{figure}

We propose \textit{stochastic ensemble value expansion} (STEVE), an extension to \textit{model-based value expansion} (MVE) proposed by~\citet{feinberg2018model}.
Both techniques use a dynamics model to compute ``rollouts'' that are used to improve the targets for temporal difference learning.
MVE rolls out a fixed length into the future, potentially accumulating model errors or increasing value estimation error along the way.
In contrast, STEVE interpolates between many different horizon lengths, favoring those whose estimates have lower uncertainty, and thus lower error.
To compute the interpolated target, we replace both the model and Q-function with ensembles, approximating the uncertainty of an estimate by computing its variance under samples from the ensemble.
%
Through these uncertainty estimates, STEVE dynamically utilizes the model rollouts only when they do not introduce significant errors. 
For illustration, Figure \ref{fig:toy} compares the sample efficiency of various algorithms on a tabular toy environment, which shows that STEVE significantly outperforms MVE and TD-learning baselines when the dynamics model is noisy. 
%
We systematically evaluate STEVE on several challenging continuous control benchmarks and demonstrate that STEVE significantly outperforms model-free baselines with an order-of-magnitude increase in sample efficiency.


\section{Background}
\label{background}
Reinforcement learning aims to learn an agent policy that maximizes the expected (discounted) sum of rewards~\citep{sutton1998reinforcement}. The agent starts at an initial state $s_0 \sim p(s_0)$, where $p(s_0)$ is the distribution of initial states of the environment. Then, the agent deterministically chooses an action $a_t$ according to its policy $\pi_\phi(s_t)$ with parameters $\phi$, deterministically transitions to a subsequent state $s_{t+1}$ according to the Markovian dynamics $T(s_t, a_t)$ of the environment, and receives a reward $r_t = r(s_t, a_t, s_{t+1})$. This generates a trajectory of states, actions, and rewards $\tau = (s_0, a_0, r_0, s_1, a_1, \ldots)$. If a trajectory reaches a terminal state, it concludes without further transitions or rewards; however, this is optional, and trajectories may instead be infinite in length. We abbreviate the trajectory by $\tau$. The goal is to maximize the expected discounted sum of rewards along sampled trajectories $J(\theta) = \mathbb{E}_{s_0} \left[ \sum_{t = 0}^\infty \gamma^t r_t \right]$ where $\gamma \in [0, 1)$ is a discount parameter.

\subsection{Value Estimation with TD-learning}
The action-value function $Q^\pi(s_0, a_0) = \sum_{t=0}^\infty \gamma^t r_t$ is a critical quantity to estimate for many learning algorithms. Using the fact that $Q^\pi(s, a)$ satisfies a recursion relation
\[ Q^\pi(s, a) = r(s, a) + \gamma (1 - d(s')) Q^\pi(s', \pi(s')), \]
where $s' = T(s, a)$ and $d(s')$ is an indicator function which returns $1$ when $s'$ is a terminal state and $0$ otherwise. We can estimate $Q^\pi(s, a)$ off-policy with collected transitions of the form $(s, a, r, s')$ sampled uniformly from a replay buffer~\citep{sutton1998reinforcement}. We approximate $Q^\pi(s, a)$ with a deep neural network, $\hat{Q}_\theta^\pi(s, a)$. We learn parameters $\theta$ to minimize the mean squared error (MSE) between Q-value estimates of states and their corresponding TD targets:
\begin{align}
\mathcal{T}^{TD}(r,s') &= r + \gamma (1 - d(s')) \hat{Q}_{\theta^-}^\pi(s', \pi(s')) \\
\label{td_update_formula}
\mathcal{L}_\theta &= \E_{(s, a, r, s')}\left[ (\hat{Q}_\theta^\pi(s, a) - \mathcal{T}^{\TD}(r,s'))^2 \right]
\end{align}

This expectation is taken with respect to transitions sampled from our replay buffer.
Note that we use an older copy of the parameters, $\theta^-$, when computing targets~\citep{mnih2013playing}.

Since we evaluate our method in a continuous action space, it is not possible to compute a policy from our Q-function by simply taking $\max_a\hat{Q}^\pi_\theta(s,a)$. Instead, we use a neural network to approximate this maximization function \cite{lillicrap2015continuous}, by learning a parameterized function $\pi_\phi$ to minimize the negative Q-value:
\begin{align}
\label{policy_update_formula}
\mathcal{L}_\phi = -\hat{Q}^\pi_\theta(s,\pi_\phi(s)).
\end{align}
%
In this work, we use DDPG as the base learning algorithm, but our technique is generally applicable to other methods that use TD objectives. 

\subsection{Model-Based Value Expansion (MVE)}
Recently, \citet{feinberg2018model} showed that a learned dynamics model can be used to improve value estimation. MVE forms TD targets by combining a short term value estimate formed by unrolling the model dynamics and a long term value estimate using the learned $\hat{Q}_{\theta^-}^\pi$ function. When the model is accurate, this reduces the bias of the targets, leading to improved performance. 

The learned dynamics model consists of three learned functions: the transition function $\hat{T}_\xi(s,a)$, which returns a
successor state $s'$; a termination function $\hat{d}_\xi(s)$, which returns the probability that $s$ is a terminal state; and the reward function $\hat{r}_\psi(s,a,s')$, which returns a scalar
reward. This model is trained to minimize 
\begin{align}
\label{model_update_formula}
 \mathcal{L}_{\xi,\psi} = \E_{(s, a, r, s')} \bigr[ ||\hat{T}_\xi(s,a) - s'||^2 + \mathbb{H}\left(d(s'), \hat{d}_\xi(\hat{T}_\xi(s, a))\right) + (\hat{r}_\psi(s, a, s') - r)^2 \bigr],
\end{align}
where the expectation is over collected transitions $(s, a, r, s')$, and $\mathbb{H}$ is the cross-entropy. In this work, we consider continuous environments; for discrete environments, the first term can be replaced by a cross-entropy loss term.

To incorporate the model into value estimation, \citet{feinberg2018model} replace the standard Q-learning target with an improved target, $\mathcal{T}_H^{\rm MVE}$, computed by rolling the learned model out for $H$ steps.
\begin{align}
\label{mve_rollout_formula}
s'_0 &= s', \qquad a'_i = \pi_\phi(s'_i), \qquad s'_i = \hat{T}_\xi(s'_{i-1}, a'_{i-1}), \qquad D^i = d(s')\prod_{j=1}^i(1-\hat{d}_\xi(s'_j))\\
\label{mve_return_formula}
\mathcal{T}_H^{\MVE}(r,s') &= r + \left(\sum_{i=1}^H D^i \gamma^i \hat{r}_\psi(s'_{i-1},a'_{i-1},s'_i)\right) + D^{H+1} \gamma^{H+1} \hat{Q}_{\theta^-}^\pi(s'_H, a'_H).
\end{align}
To use this target, we substitute $\mathcal{T}_H^{\MVE}$ in place of $\mathcal{T}^{\TD}$ when training $\theta$ using Equation~\ref{td_update_formula}.\footnote{This formulation is a minor generalization of the original MVE objective in that we additionally model the reward function and termination function; \citet{feinberg2018model} consider ``fully observable'' environments in which the reward function and termination condition were known, deterministic functions of the observations. Because we use a function approximator for the termination condition, we compute the accumulated probability of termination, $D^i$, at every timestep, and use this value to discount future returns.}
Note that when $H = 0$, MVE reduces to TD-learning (i.e., $\mathcal{T}^{\TD} = \mathcal{T}_0^{\MVE}$). 

When the model is perfect and the learned Q-function has similar bias on all states and actions, \citet{feinberg2018model} show that the MVE target with rollout horizon $H$ will decrease the target error by a factor of $\gamma^{2H}$. Errors in the learned model can lead to worse targets, so in practice, we must tune $H$ to balance between the errors in the model and the $Q$-function estimates. An additional challenge is that the bias in the learned Q-function is not uniform across states and actions~\citep{feinberg2018model}. In particular, they find that the bias in the Q-function on states sampled from the replay buffer is lower than when the Q-function is evaluated on states generated from model rollouts.
They term this the \textit{distribution mismatch problem} and propose the \textit{TD-k trick} as a solution; see Appendix \ref{appendix:tdk} for further discussion of this trick.

While the results of \citet{feinberg2018model} are promising, they rely on task-specific tuning of the rollout horizon $H$. This sensitivity arises from the difficulty of modeling the transition dynamics and the $Q$-function, which are task-specific and may change throughout training as the policy explores different parts of the state space. Complex environments require much smaller rollout horizon $H$, which limits the effectiveness of the approach (e.g., \citet{feinberg2018model} used $H = 10$ for HalfCheetah-v1, but had to reduce to $H = 3$ on Walker2d-v1). Motivated by this limitation, we propose an approach that balances model error and Q-function error by dynamically adjusting the rollout horizon.

\section{Stochastic Ensemble Value Expansion}
\label{method}

From a single rollout of $H$ timesteps, we can compute $H+1$ distinct \textit{candidate targets} by considering rollouts of various horizon lengths:
$\mathcal{T}_0^{\MVE}$,$\mathcal{T}_1^{\MVE}$,$\mathcal{T}_2^{\MVE}$,$...$,$\mathcal{T}_H^{\MVE}$. 
Standard TD learning uses $\mathcal{T}_0^{\MVE}$ as the target, while MVE uses $\mathcal{T}_H^{\MVE}$ as the target.
We propose interpolating all of the candidate targets to produce a target which is better than any individual.
Conventionally, one could average the candidate targets, or weight the candidate targets in an exponentially-decaying fashion, similar to TD($\lambda$)~\cite{sutton1998reinforcement}. However, we show that we can do still better by weighting the candidate targets in a way that balances errors in the learned $Q$-function and errors from longer model rollouts. STEVE provides a computationally-tractable and theoretically-motivated algorithm for choosing these weights. We describe the algorithm for STEVE in Section~\ref{algo-sec}, and justify it in Section~\ref{justification}.

\subsection{Algorithm}
\label{algo-sec}

To estimate uncertainty in our learned estimators, we maintain an ensemble of parameters for our Q-function, reward function, and model: 
$\boldsymbol{\theta} = \{\theta_1, ..., \theta_L \}$, $\boldsymbol{\psi} = \{\psi_1, ..., \psi_N\}$, and $\boldsymbol{\xi} = \{\xi_1, ..., \xi_M \}$, respectively. Each parameterization is initialized independently and trained on different subsets of the data in each
minibatch.

\begin{figure}[t]
\centering
    \includegraphics[width=\textwidth]{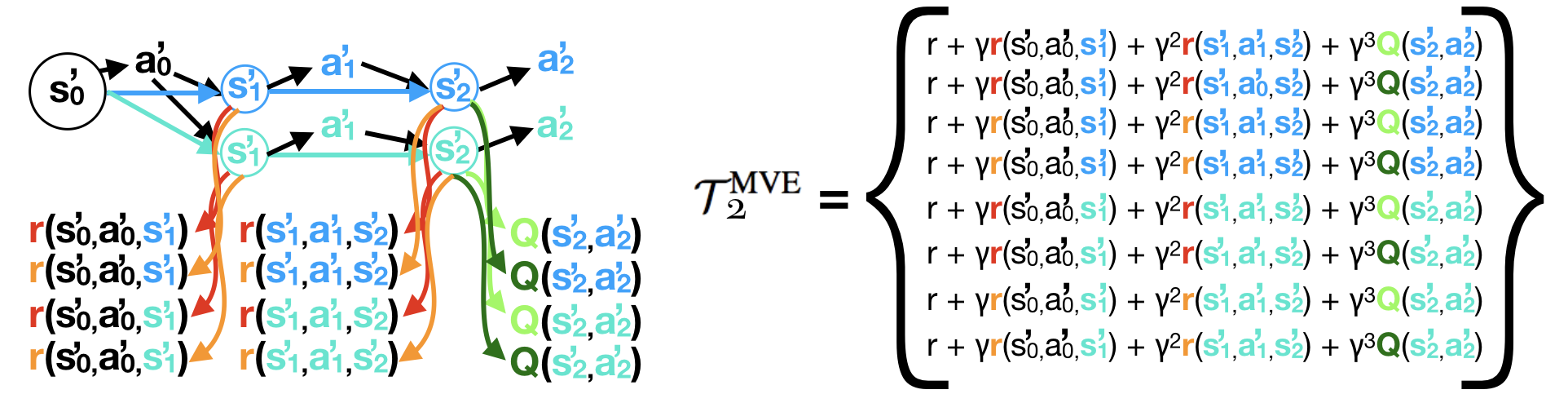}  
\caption{Visualization of how the set of possible values for each candidate target is computed, shown for a length-two rollout with $M,N,L=2$. Colors correspond to ensemble members. Best viewed in color.}
\label{fig:target-illustration}
\end{figure}

We roll out an $H$ step trajectory with each of the $M$ models, $\tau^{\xi_1},...,\tau^{\xi_M}$. Each trajectory consists of $H+1$ states, $\tau^{\xi_m}_0,...,\tau^{\xi_m}_H$, which correspond to $s'_0,...,s'_H$ in Equation~\ref{mve_rollout_formula} with the transition function parameterized by $\xi_m$. Similarly, we use the $N$ reward functions and $L$ Q-functions to evaluate Equation~\ref{mve_return_formula} for each $\tau^{\xi_m}$ at every rollout-length $0 \leq i \leq H$. This gives us $M \cdot N \cdot L$ different values of $\mathcal{T}^{\MVE}_i$ for each rollout-length $i$. See Figure \ref{fig:target-illustration} for a visualization of this process.

Using these values, we can compute the empirical mean $\mathcal{T}^{\mu}_{i}$ and variance $\mathcal{T}^{\sigma^2}_{i}$ for each partial rollout of length $i$. In order to form a single target, we use an inverse variance weighting of the means:
\begin{align}
\label{steve_formula}
\mathcal{T}_H^{\STEVE}(r, s') &= \sum_{i=0}^H \frac{\tilde{w}_i}{\sum_j \tilde{w}_j }  \mathcal{T}_{i}^{\mu},\qquad \tilde{w}_i^{-1} = \mathcal{T}_{i}^{\sigma^2}
\end{align}
To learn a value function with STEVE, we substitute in $\mathcal{T}_H^{\STEVE}$ in place of $\mathcal{T}^{\TD}$ when training $\theta$ using Equation~\ref{td_update_formula}.

\subsection{Derivation}
\label{justification}

We wish to find weights $w_i$, where $\sum_i w_i = 1$ that minimize the mean-squared error between the weighted-average of candidate targets $\mathcal{T}_0^{\MVE}$,$\mathcal{T}_1^{\MVE}$,$\mathcal{T}_2^{\MVE}$,...,$\mathcal{T}_H^{\MVE}$ and the true Q-value. 
\begin{align*}
    \mathop{\E} \left[ \left( \sum_{i=0}^H w_i \mathcal{T}_{i}^{\MVE} - Q^\pi(s, a) \right)^2 \right] &= \Bias\left(\sum_i w_i \mathcal{T}_{i}^{\MVE} \right)^2 + \Var\left( \sum_i w_i \mathcal{T}_{i}^{\MVE} \right) \\
    &\approx \Bias\left(\sum_i w_i \mathcal{T}_{i}^{\MVE} \right)^2 + \sum_i w_i^2 \Var(\mathcal{T}_i^{\MVE}),
\end{align*}
where the expectation considers the candidate targets as random variables conditioned on the collected data and minibatch sampling noise, and the approximation is due to assuming the candidate targets are independent\footnote{Initial experiments suggested that omitting the covariance cross terms provided significant computational speedups at the cost of a slight performance degradation. As a result, we omitted the terms in the rest of the experiments.}. 

Our goal is to minimize this with respect to $w_i$. We can estimate the variance terms using empirical variance estimates from the ensemble. Unfortunately, we could not devise a reliable estimator for the bias terms, and this is a limitation of our approach and an area for future work. In this work, we ignore the bias terms and minimize the weighted sum of variances
\[ \sum_i w_i^2 \Var(\mathcal{T}_i^{\MVE}). \]
With this approximation, which is equivalent to in inverse-variance weighting \cite{fleiss1993review}, we achieve state-of-the-art results. 
Setting each $w_i$ equal to $\frac{1}{\Var(\mathcal{T}_i^{\MVE})}$ and normalizing yields the formula for $\mathcal{T}_H^{\STEVE}$ given in Equation~\ref{steve_formula}.

\subsection{Note on ensembles}
This technique for calculating uncertainty estimates is applicable to any family of models from which we can sample. For example, we could train a Bayesian neural network for each model~\cite{mackay1992practical}, or use dropout as a Bayesian approximation by resampling the dropout masks each time we wish to sample a new model~\cite{gal2016dropout}. These options could potentially give better diversity of various samples from the family, and thus better uncertainty estimates; exploring them further is a promising direction for future work. However, we found that these methods degraded the accuracy of the base models. An ensemble is far easier to train, and so we focus on that in this work.
This is a common choice, as the use of ensembles in the context of uncertainty estimations for deep reinforcement learning has seen wide adoption in the literature. It was first proposed by~\citet{osband2016deep} as a technique to improve exploration, and subsequent work showed that this approach gives a good estimate of the uncertainty of both value functions~\cite{kalweit2017uncertainty} and models~\cite{kurutach2018modelensemble}.
\section{Experiments}
\label{experiments}

\subsection{Implementation}
We use DDPG \cite{lillicrap2015continuous} as our baseline model-free algorithm. We train two deep feedforward neural networks,
a Q-function network $\hat{Q}^\pi_\theta(s,a)$ and a policy network $\pi_\phi(s)$, by minimizing the loss functions given in Equations \ref{td_update_formula} and \ref{policy_update_formula}. We also train another three deep feedforward networks to represent our world model, corresponding to function approximators for the transition $\hat{T}_\xi(s,a)$, termination $\hat{d}_\xi(t \mid s)$, and reward $\hat{r}_\psi(s,a,s')$, and minimize the loss function given in Equation \ref{model_update_formula}.

\begin{figure}[t]
\centering
\begin{subfigure}[t]{.32\textwidth}
\centering
\includegraphics[width=\textwidth]{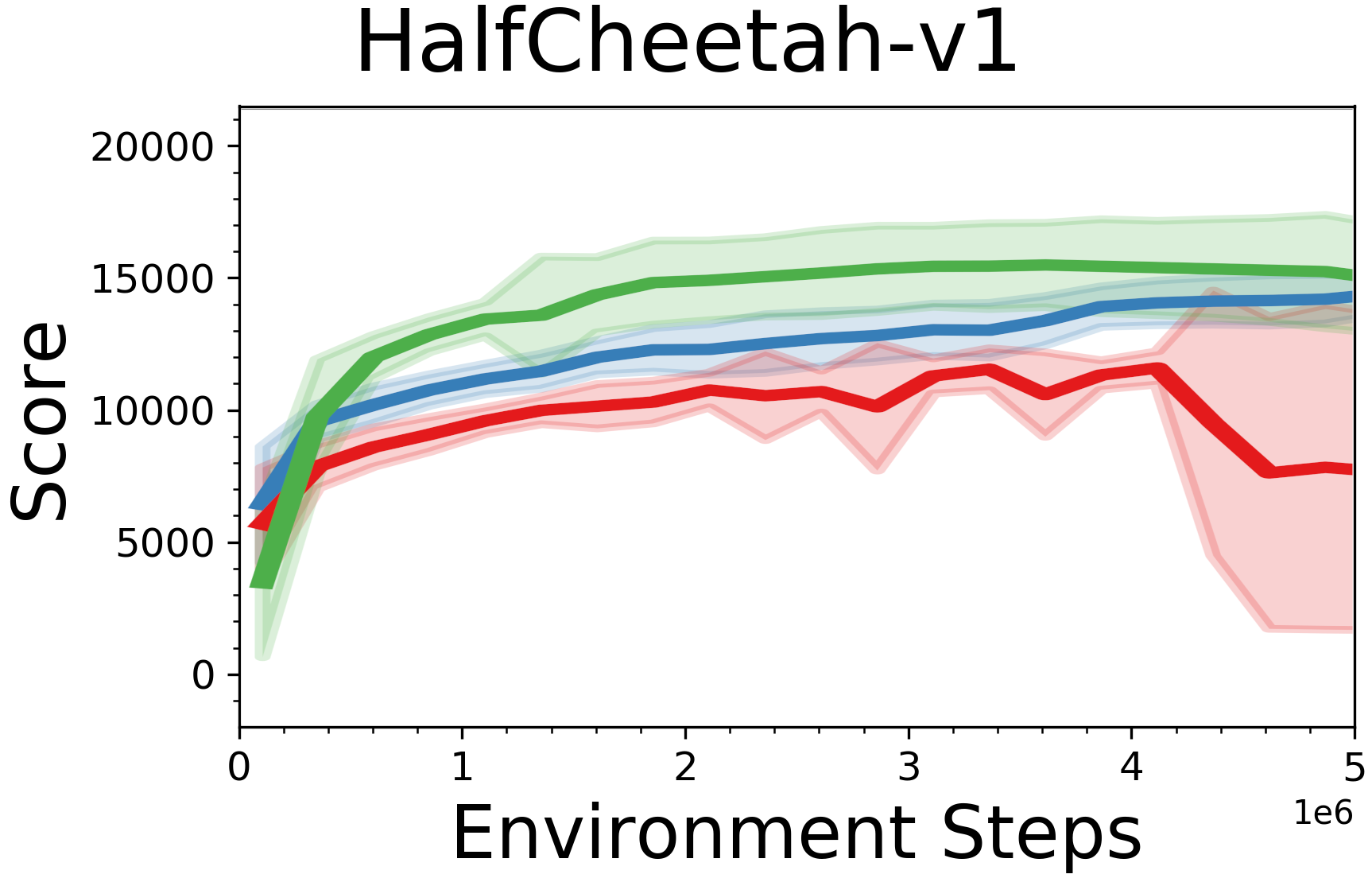}
\end{subfigure}\hfill%
\begin{subfigure}[t]{.32\textwidth}
\centering
\includegraphics[width=\textwidth]{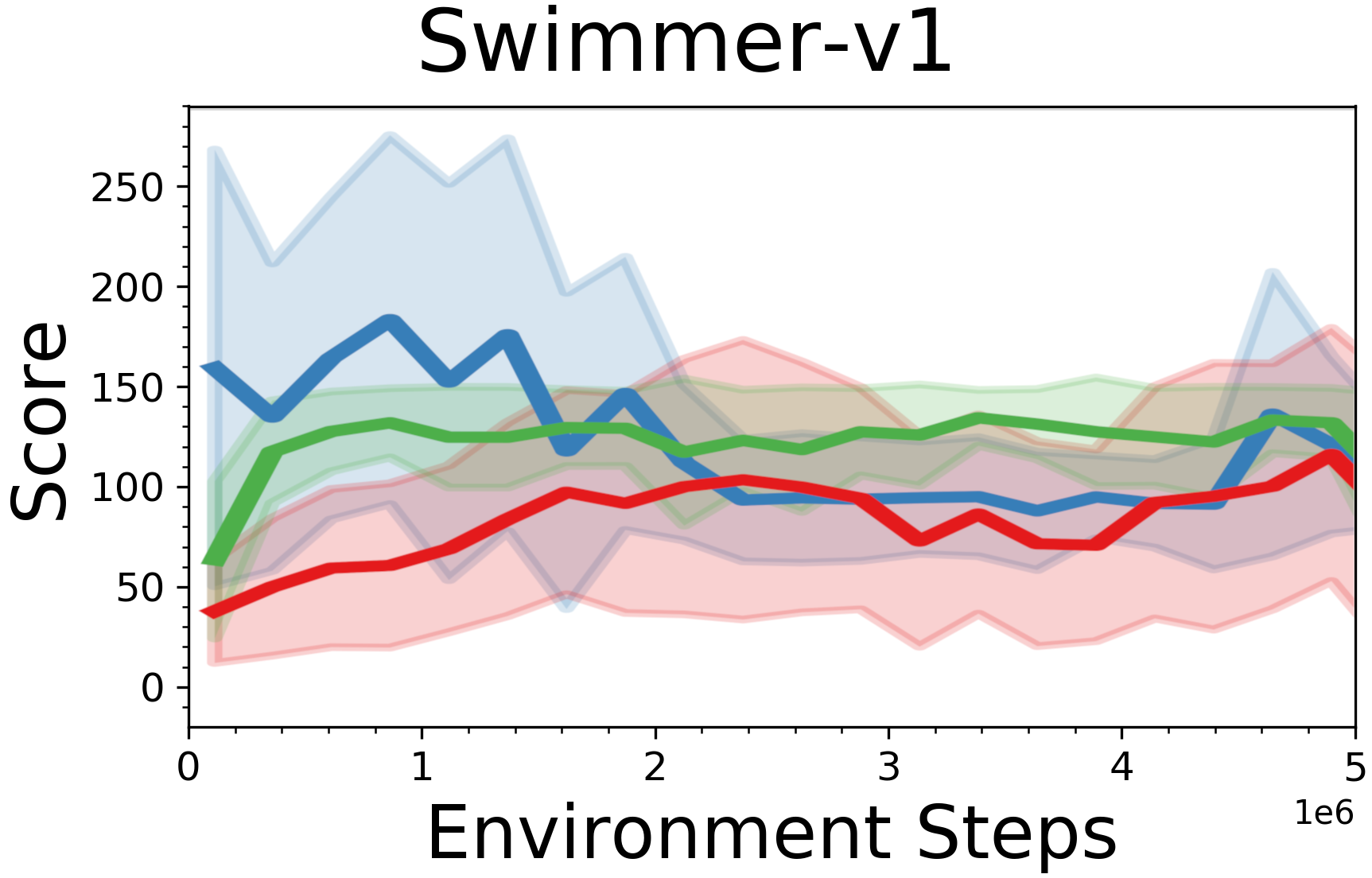}
\end{subfigure}\hfill%
\begin{subfigure}[t]{.32\textwidth}
\centering
\includegraphics[width=\textwidth]{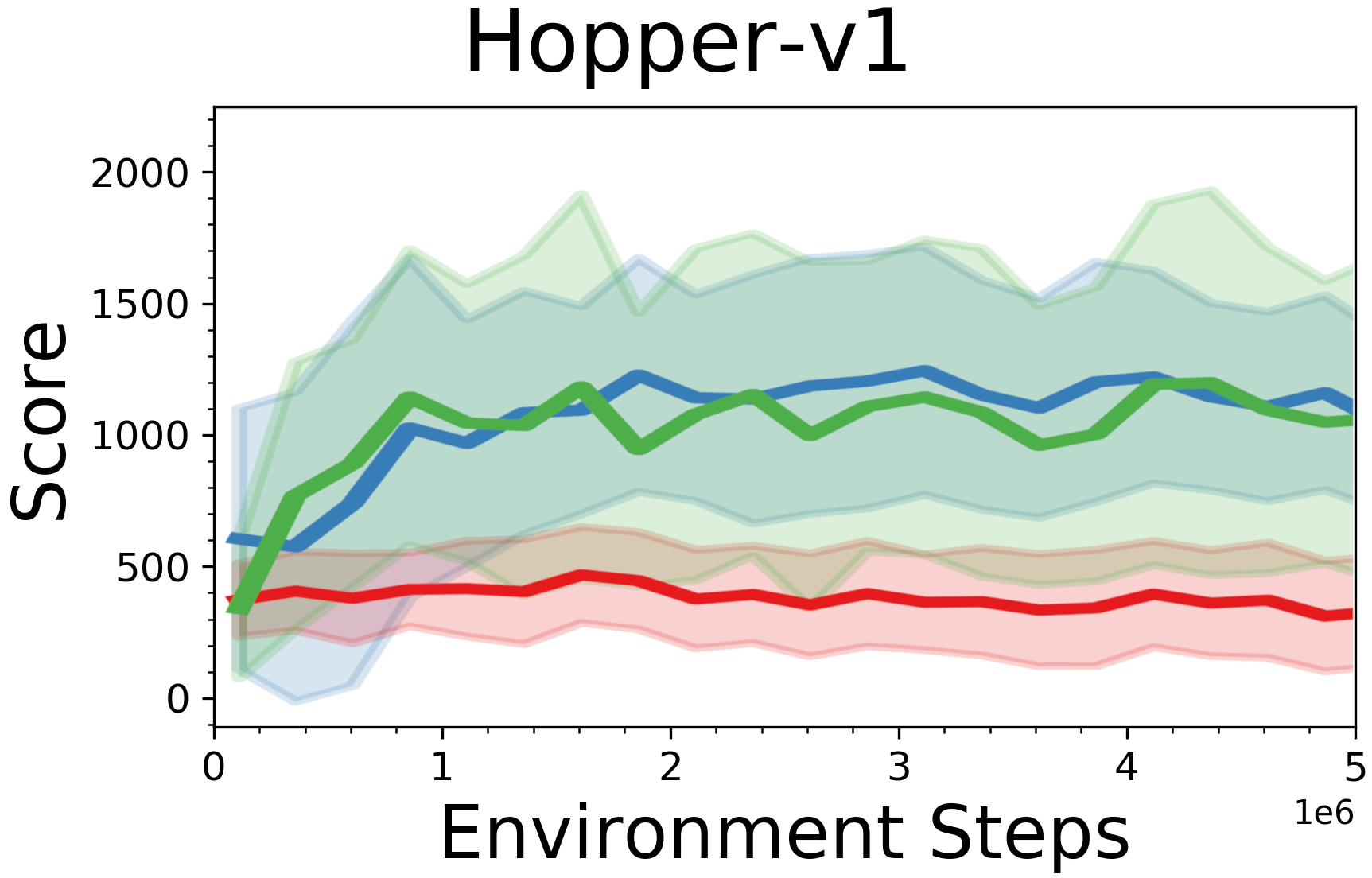}
\end{subfigure}\hfill%
\\
\begin{subfigure}[t]{.32\textwidth}
\centering
\includegraphics[width=\textwidth]{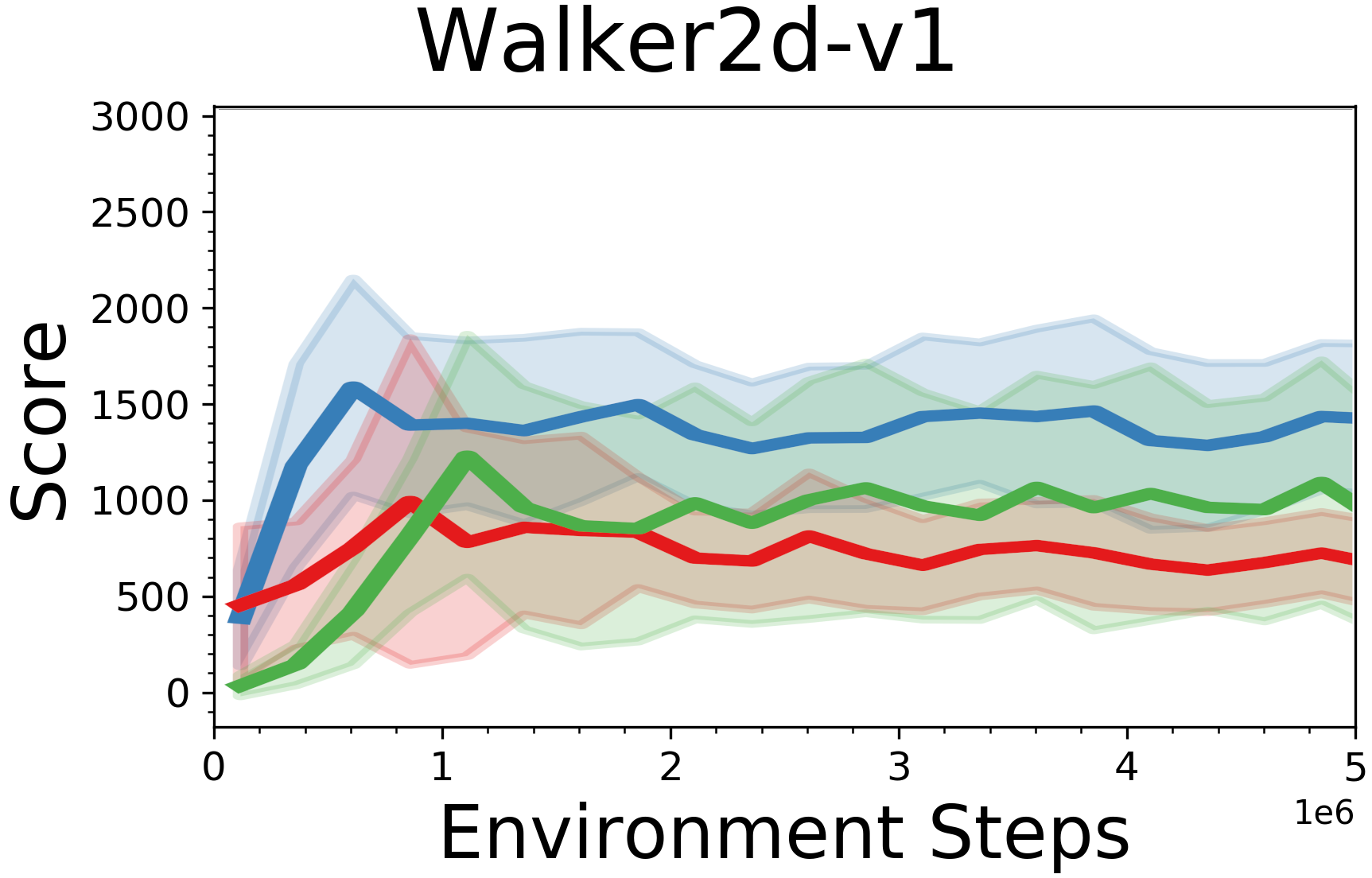}
\end{subfigure}
\hspace{0.03in}
\begin{subfigure}[t]{.32\textwidth}
\centering
\includegraphics[width=\textwidth]{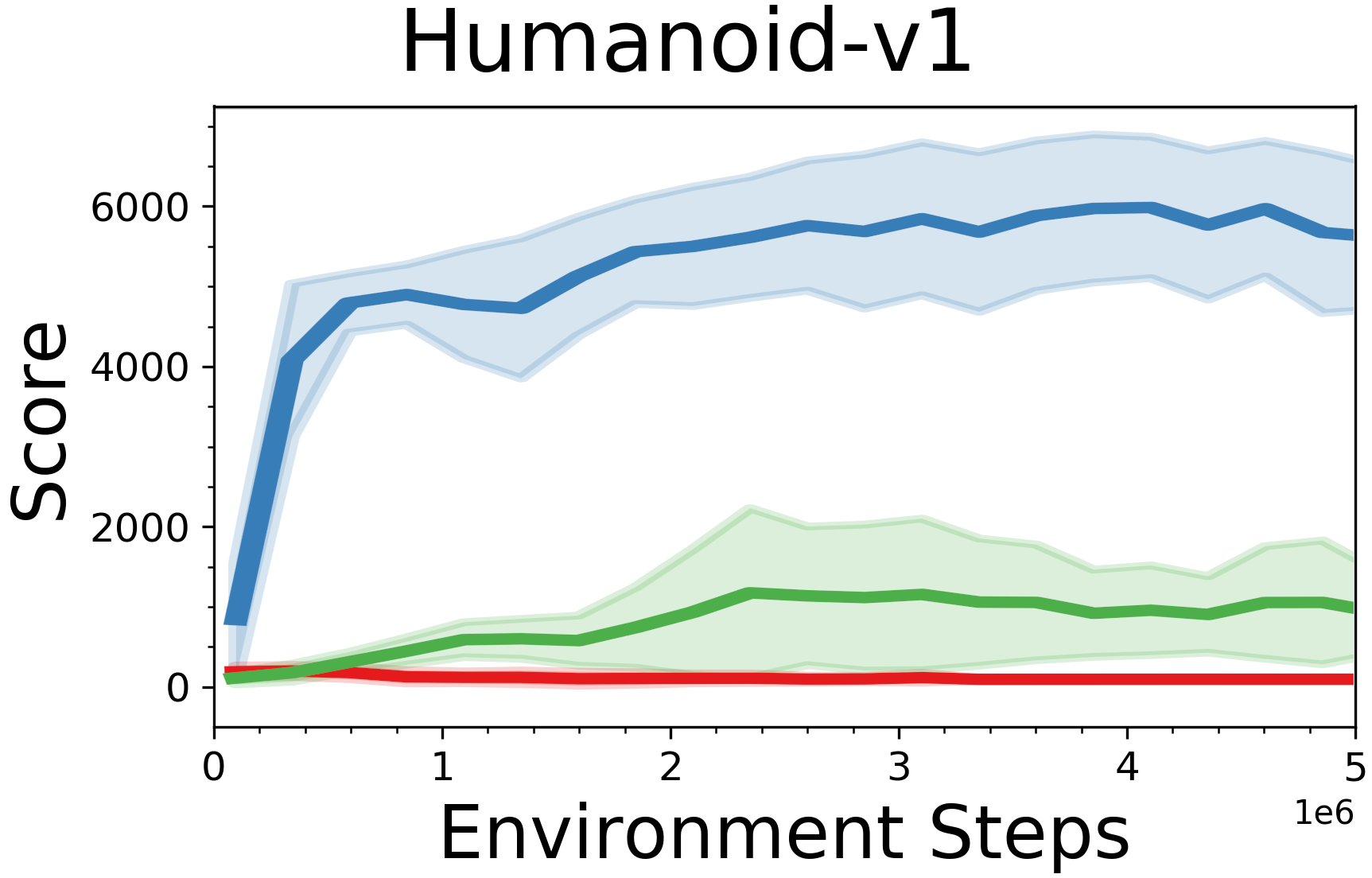}
\end{subfigure}\hfill%
\begin{subfigure}[t]{.32\textwidth}
\centering
\includegraphics[width=\textwidth]{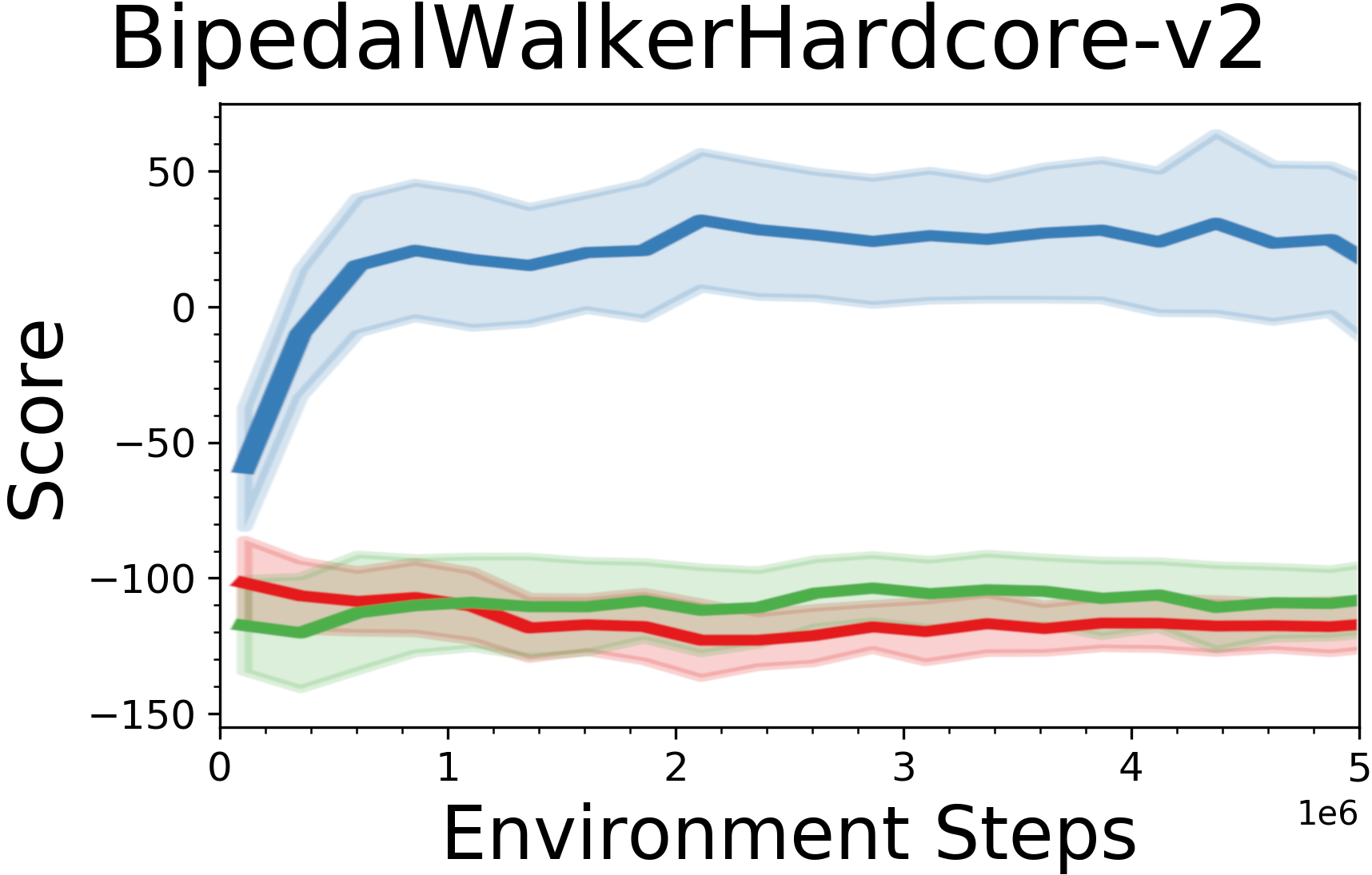}
\end{subfigure}\hfill%
\\
\hspace{-0.15in}
\begin{subfigure}[t]{.32\textwidth}
\centering
\includegraphics[width=\textwidth]{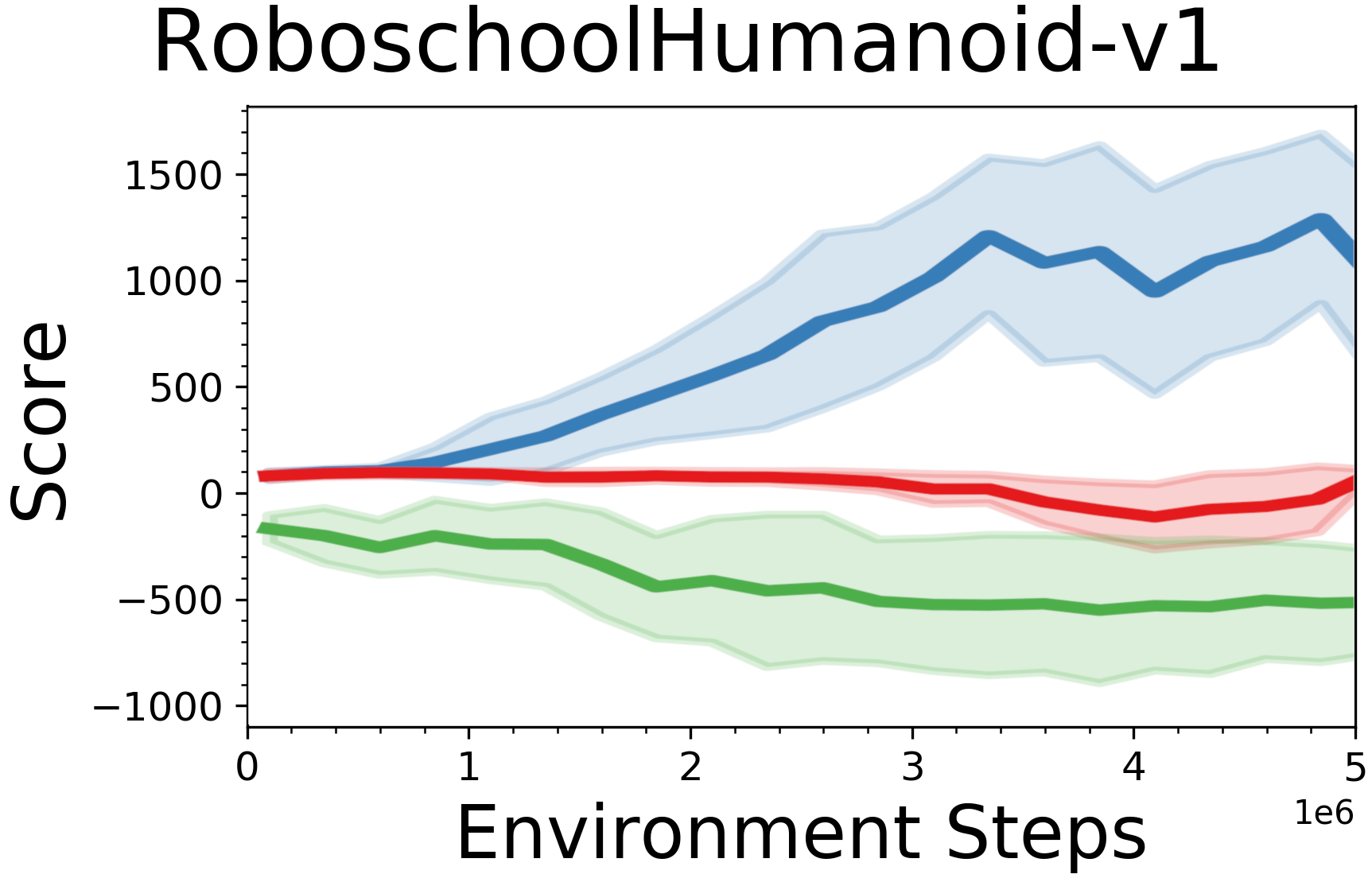}
\end{subfigure}
\hspace{0.05in}
\begin{subfigure}[t]{.32\textwidth}
\centering
\includegraphics[width=\textwidth]{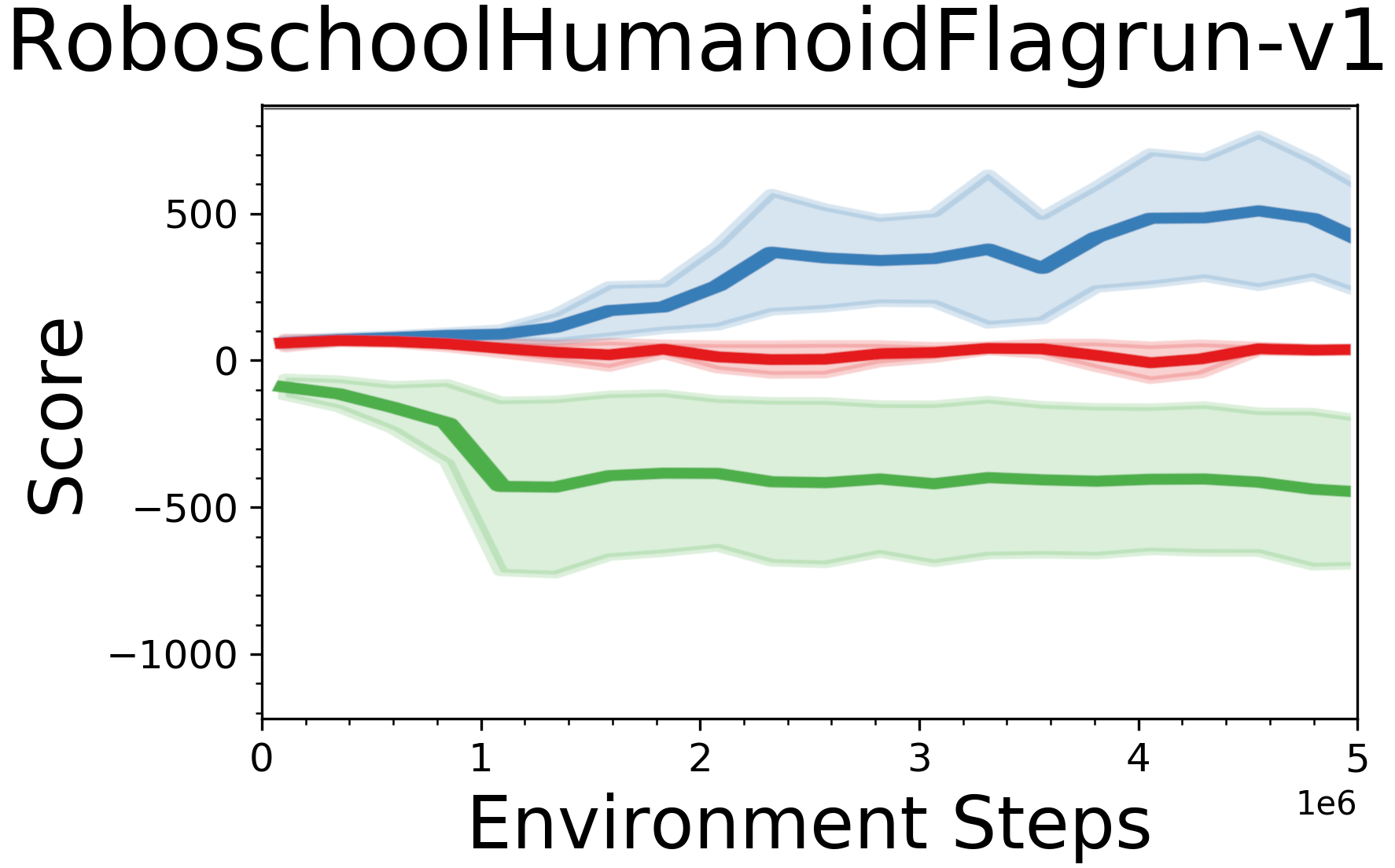}
\end{subfigure}
\begin{subfigure}[t]{.32\textwidth}
\centering
\includegraphics[width=\textwidth]{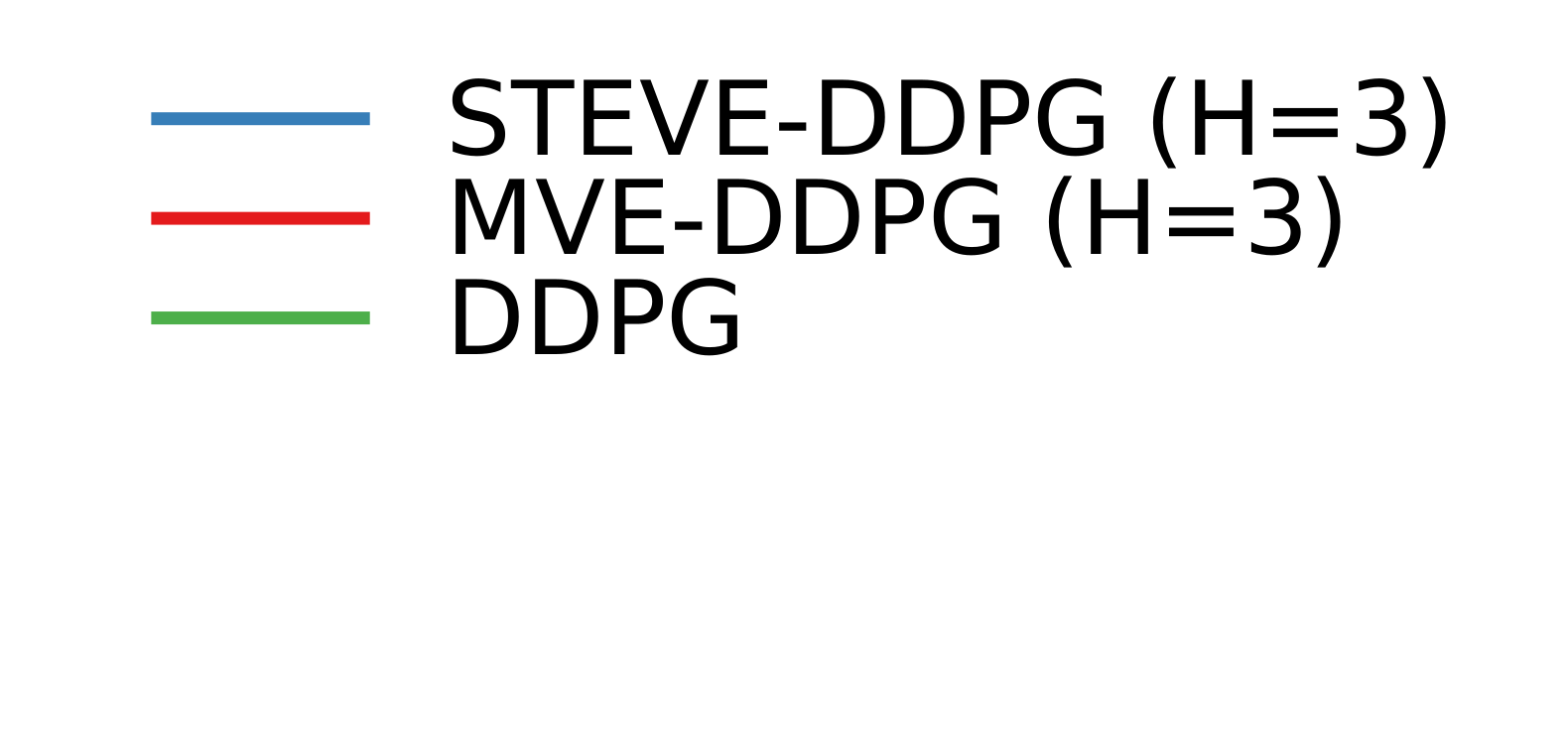}
\end{subfigure}
\caption{Learning curves comparing sample efficiency of our method to both model-free and model-based baselines.
Each experiment was run four times.}
\label{fig:sample-efficiency}
\end{figure}

When collecting rollouts for evaluation, we simply take the action selected by the
policy, $\pi_\phi(s)$, at every state $s$. (Note that only the policy is required at test-time, not the ensembles of Q-functions, dynamics models, or reward models.) Each run was evaluated after every 500 updates by computing the mean total episode reward (referred to as score) across many environment restarts. To produce the lines in Figures \ref{fig:sample-efficiency}, \ref{fig:ablation}, and \ref{fig:wall-clock-time}, these evaluation results were downsampled by splitting the domain into non-overlapping regions and computing the mean score within each region across several runs. The shaded area shows one standard deviation of scores in the region as defined above.

When collecting rollouts for our replay buffer, we do $\epsilon$-greedy exploration:
with probability $\epsilon$, we select a random action by adding Gaussian noise to the pre-tanh
policy action.

All algorithms were implemented in Tensorflow~\citep{abadi2016tensorflow}. 
We use a distributed implementation to parallelize computation. In the style of
ApeX~\cite{horgan2018distributed}, IMPALA~\cite{espeholt2018impala}, and D4PG~\cite{barth-maron2018distributional},
we use a centralized learner with several agents operating in parallel.
Each agent periodically loads the most recent policy, interacts with the environment, and sends
its observations to the central learner. The learner stores received frames in a replay buffer,
and continuously loads batches of frames from this buffer to use as training data for a model
update. In the algorithms with a model-based component, there are two learners: a policy-learner
and a model-learner. In these cases, the policy-learner periodically reloads the latest copy of
the model.

All baselines reported in this section were re-implementations of existing methods. This allowed us to ensure that the various methods compared were consistent with one another, and that the differences reported are fully attributable to the independent variables in question. Our baselines are competitive with state-of-the-art implementations of these algorithms~\cite{haarnoja2018soft,feinberg2018model}. All MVE experiments utilize the TD-k trick. For hyperparameters and additional implementation details, please see Appendix~\ref{implement}.\footnote{Our code is available open-source at: \url{https://github.com/tensorflow/models/tree/master/research/steve}}

\subsection{Comparison of Performance}
\label{experiment:performance}

We evaluated STEVE on a variety of continuous control tasks~\citep{brockman2016openai,roboschool}; we plot learning curves in Figure~\ref{fig:sample-efficiency}. We found that STEVE yields significant improvements in both performance and sample efficiency across a wide range of environments.
Importantly, the gains are most substantial in the complex environments. On the most challenging environments: Humanoid-v1, RoboschoolHumanoid-v1, RoboschoolHumanoidFlagrun-v1, and BipedalWalkerHardcore-v2, STEVE is the only algorithm to show significant learning within 5M frames.


\subsection{Ablation Study}
\label{experiment:ablation}

\begin{wrapfigure}{r}{0.5\textwidth}
\vspace{-0.25in}
  \begin{center}
    \includegraphics[width=0.5\textwidth]{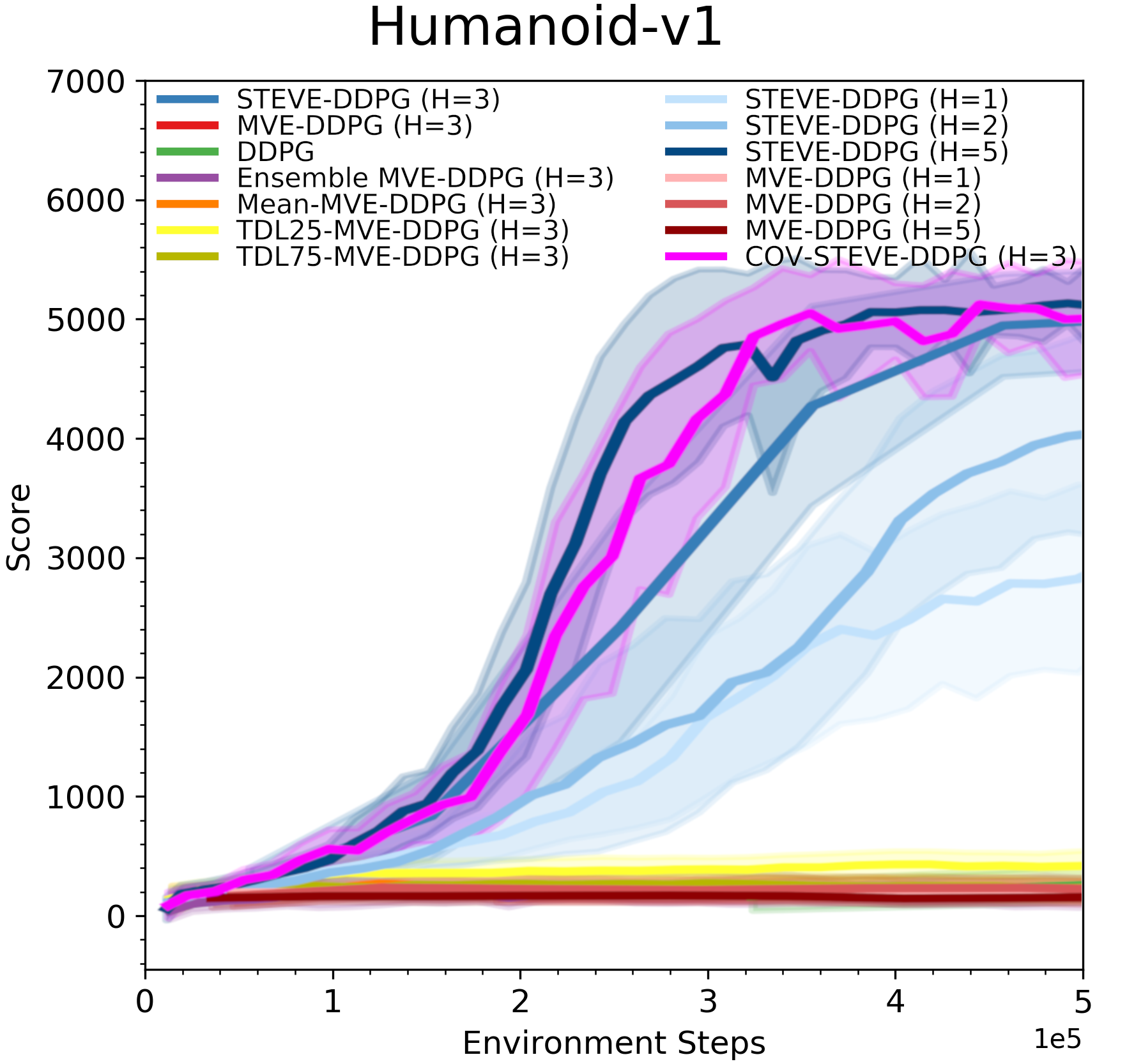}  
    \end{center}
\caption{Ablation experiments on variation of methods.
Each experiment was run twice.}
\label{fig:ablation}
\end{wrapfigure}
In order to verify that STEVE's gains in sample efficiency are due to the reweighting, and not simply due to the additional parameters of the ensembles of its components, we examine several ablations. Ensemble MVE is the regular MVE algorithm, but the model and Q-functions are replaced with with ensembles. Mean-MVE uses the exact same architecture as STEVE, but uses a simple uniform weighting instead of the uncertainty-aware reweighting scheme. Similarly, TDL25 and TDL75 correspond to TD($\lambda$) reweighting schemes with $\lambda=0.25$ and $\lambda=0.75$, respectively. COV-STEVE is a version of STEVE which includes the covariances between candidate targets when computing the weights (see Section~\ref{justification}). We also investigate the effect of the horizon parameter on the performance of both STEVE and MVE. These results are shown in Figure~\ref{fig:ablation}.

%
All of these variants show the same trend: fast initial gains, which quickly taper off and are overtaken by the baseline. STEVE is the only variant to converge faster and higher than the baseline; this provides strong evidence that the gains come specifically from the uncertainty-aware reweighting of targets. Additionally, we find that increasing the rollout horizon increases the sample efficiency of STEVE, even though the dynamics model for Humanoid-v1 has high error.

\begin{figure}[t]
\centering
\begin{subfigure}[t]{\textwidth}
\centering
\includegraphics[width=0.56\textwidth,trim={7cm 0 7cm 0},clip]{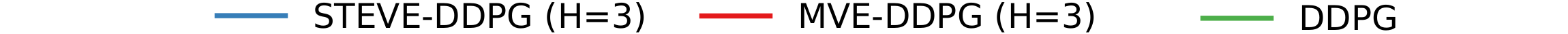}
\end{subfigure}
\par\medskip
\begin{subfigure}[t]{.3\textwidth}
\centering
\includegraphics[width=\textwidth]{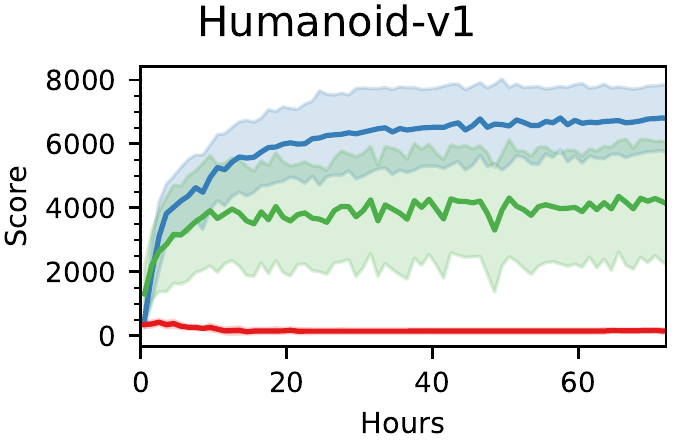}
\end{subfigure}%
\hspace{0.25in}
\begin{subfigure}[t]{.3\textwidth}
\centering
\includegraphics[width=\textwidth]{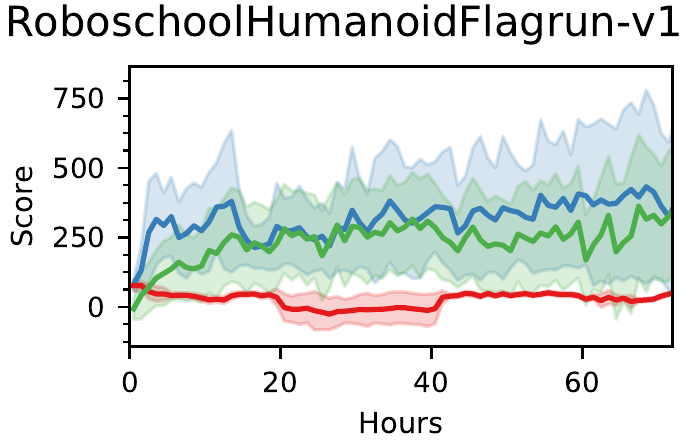}
\end{subfigure}
\caption{Comparison of wall-clock time between our method and baselines.
Each experiment was run three times.}
\label{fig:wall-clock-time}
\vspace*{-0.15in}
\end{figure}

\subsection{Wall-Clock Comparison}
\label{experiment:wallclock}
In the previous experiments, we synchronized data collection, policy updates, and model updates. However, when we run these steps asynchronously, we can reduce the wall-clock time at the risk of instability. To evaluate this configuration, we compare DDPG, MVE-DDPG, and STEVE-DPPG on Humanoid-v1 and RoboschoolHumanoidFlagrun-v1. Both were
trained on a P100 GPU and had 8 CPUs collecting data; STEVE-DPPG additionally used a second P100 to learn a model in parallel. We plot reward as a function of wall-clock time for these tasks in Figure~\ref{fig:wall-clock-time}. STEVE-DDPG learns more quickly on both tasks, and it achieves a higher reward than DDPG and MVE-DDPG on Humanoid-v1 and performs comparably to DDPG on RoboschoolHumanoidFlagrun-v1. Moreover, in future work, STEVE could be accelerated by parallelizing training of each component of the ensemble.

\section{Discussion}
\begin{figure}[b]
\vspace*{-0.1in}
\centering
\begin{subfigure}[t]{.24\textwidth}
\centering
\includegraphics[width=\textwidth]{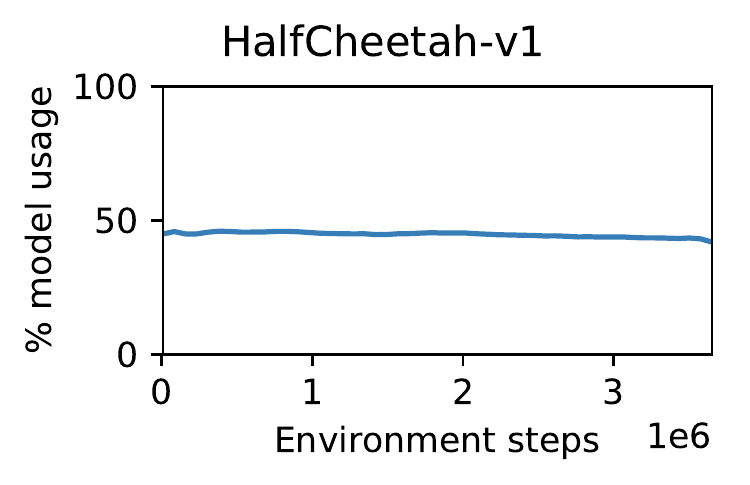}
\end{subfigure}\hfill%
\begin{subfigure}[t]{.24\textwidth}
\centering
\includegraphics[width=\textwidth]{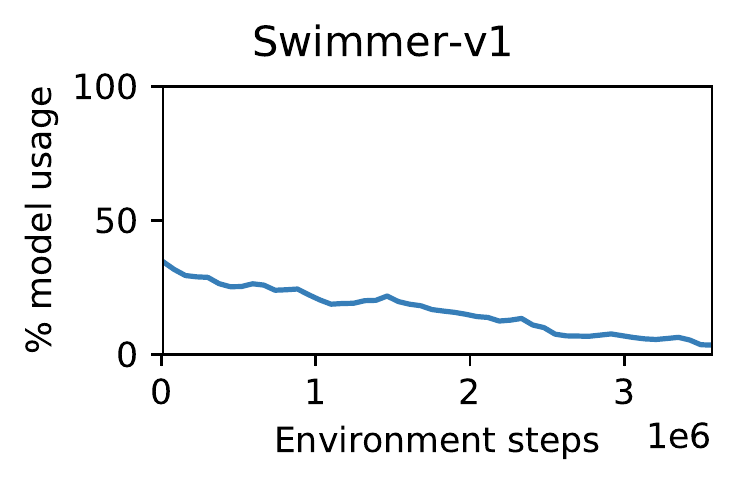}
\end{subfigure}\hfill%
\begin{subfigure}[t]{.24\textwidth}
\centering
\includegraphics[width=\textwidth]{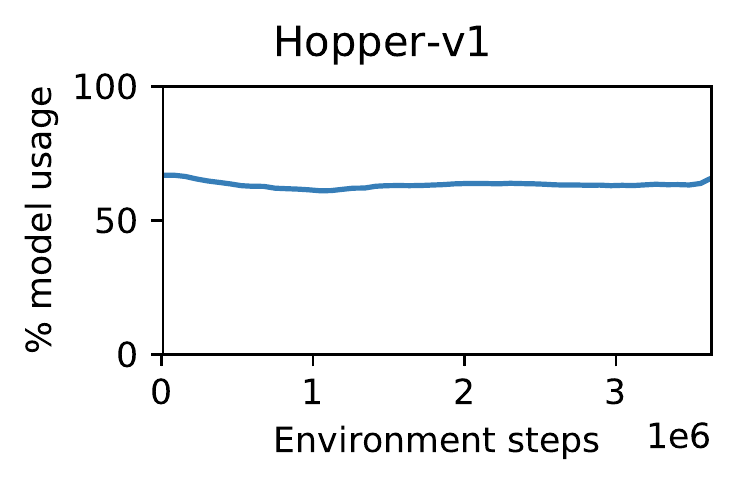}
\end{subfigure}\hfill%
\begin{subfigure}[t]{.24\textwidth}
\centering
\includegraphics[width=\textwidth]{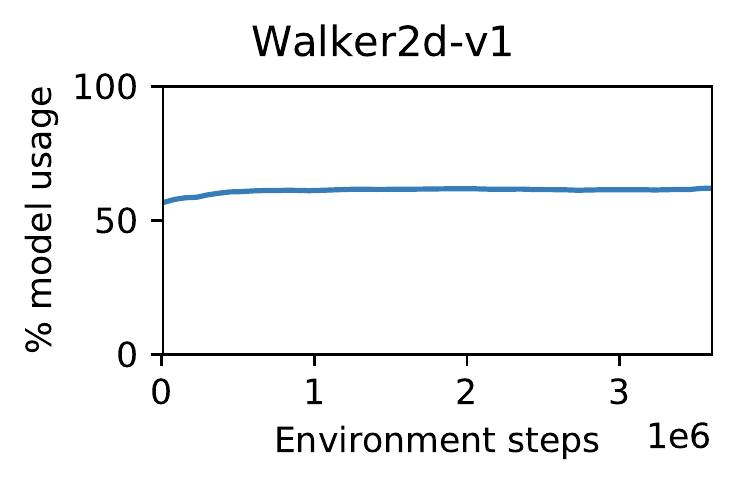}
\end{subfigure}
\\
\begin{subfigure}[t]{.24\textwidth}
\centering
\includegraphics[width=\textwidth]{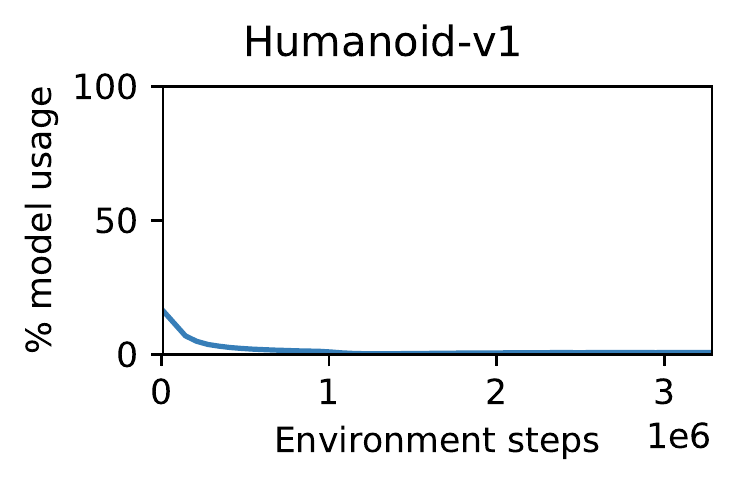}
\end{subfigure}\hfill%
\begin{subfigure}[t]{.24\textwidth}
\centering
\includegraphics[width=\textwidth]{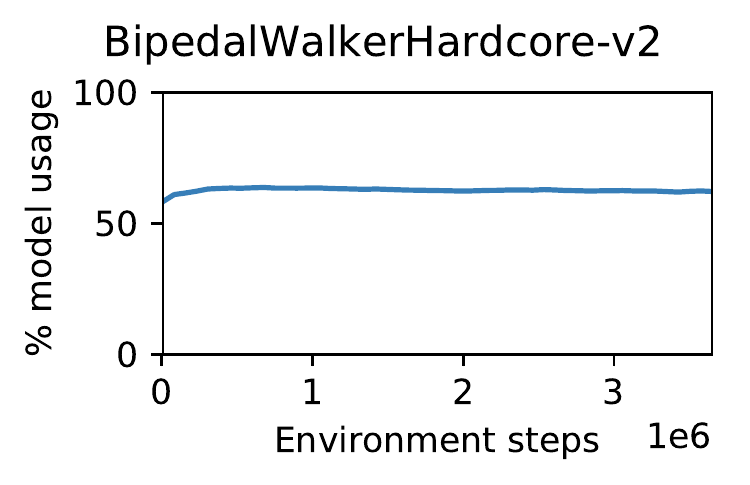}
\end{subfigure}\hfill%
\begin{subfigure}[t]{.24\textwidth}
\centering
\includegraphics[width=\textwidth]{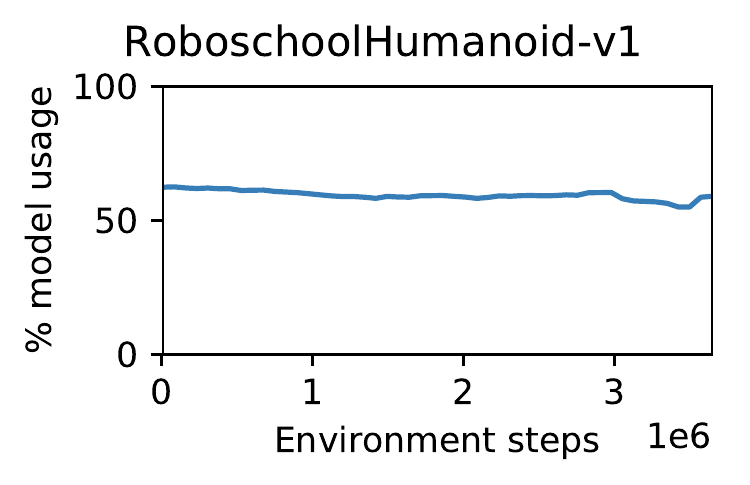}
\end{subfigure}\hfill%
\begin{subfigure}[t]{.24\textwidth}
\centering
\includegraphics[width=\textwidth]{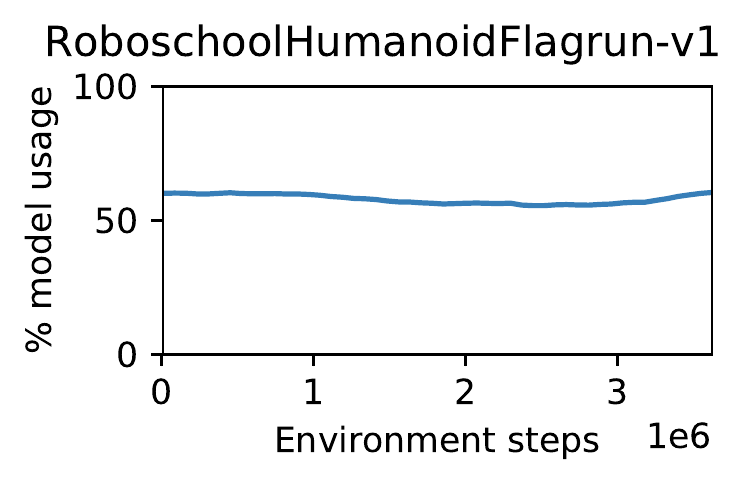}
\end{subfigure}
\caption{Average model usage for STEVE on each environment.
The y-axis represents the amount of probability mass assigned to weights that were not $w_0$, i.e. the probability mass assigned to candidate targets that include at least one step of model rollout.}
\label{fig:model-usage}
\end{figure}

Our primary experiments (Section \ref{experiment:performance}) show that STEVE greatly increases sample efficiency relative to baselines, matching or out-performing both MVE-DDPG and DDPG baselines on every task. STEVE also outperforms other recently-published results on these tasks in terms of sample efficiency~\cite{gu2016q,haarnoja2018soft,schulman2017proximal}. Our ablation studies (Section \ref{experiment:ablation}) support the hypothesis that the increased performance is due to the uncertainty-dependent reweighting of targets, as well as demonstrate that the performance of STEVE consistently increases with longer horizon lengths, even in complex environments. Finally, our wall-clock experiments (Section \ref{experiment:wallclock}) demonstrate that in spite of the additional computation per epoch, the gains in sample efficiency are enough that it is competitive with model-free algorithms in terms of wall-clock time. The speed gains associated with improved sample efficiency will only be exacerbated as samples become more expensive to collect, making STEVE a promising choice for applications involving real-world interaction.

Given that the improvements stem from the dynamic reweighting between horizon lengths, it may be interesting to examine the choices that the model makes about which candidate targets to favor most heavily. In Figure \ref{fig:model-usage}, we plot the average model usage over the course of training. Intriguingly, most of the lines seem to remain stable at around 50\% usage, with two notable exceptions: Humanoid-v1, the most complex environment tested (with an observation-space of size 376); and Swimmer-v1, the least complex environment tested (with an observation-space of size 8). This supports the hypothesis that STEVE is trading off between Q-function bias and model bias; it chooses to ignore the model almost immediately when the environment is too complex to learn, and gradually ignores the model as the Q-function improves if an optimal environment model is learned quickly.
\section{Related Work}
\label{related_work}

\citet{sutton1998reinforcement} describe TD$(\lambda)$, a family of Q-learning variants in which targets from multiple timesteps are merged via exponentially decay. STEVE is similar in that it is also computing a weighted average between targets. However, our approach is significantly more powerful because it adapts the weights to the specific characteristics of each individual rollout, rather than being constant between examples and throughout training. Our approach can be thought of as a generalization of TD($\lambda$), in that the two approaches are equivalent in the specific case where the overall uncertainty grows exponentially at rate $\lambda$ at every timestep.

\citet{munos2016safe} propose Retrace($\lambda$), a low-variance method for off-policy Q-learning. Retrace($\lambda$) is an off-policy correction method, so it learns from n-step off-policy data by multiplying each term of the loss by a correction coefficient, the \textit{trace}, in order to re-weight the data distribution to look more like the on-policy distribution. Specifically, at each timestep, Retrace($\lambda$) updates the coefficient for that term by multiplying it by $\lambda \texttt{min}(1, \frac{\pi(a_s \mid x_s)}{\mu(a_s \mid x_s)})$. Similarly to TD$(\lambda)$, the $\lambda$ parameter corresponds to an exponential decay of the weighting of potential targets. STEVE approximates this weighting in a more complex way, and additionally learns a predictive model of the environment (under which on-policy rollouts are possible) instead of using off-policy correction terms to re-weight real off-policy rollouts.

\citet{heess2015learning} describe \textit{stochastic value gradient} (SVG) methods, which are a general family of hybrid model-based/model-free control algorithms. By re-parameterizing distributions to separate out the noise, SVG is able to learn stochastic continuous control policies in stochastic environments. STEVE currently operates only with deterministic policies and environments, but this is a promising direction for future work.

\citet{kurutach2018modelensemble} propose \textit{model-ensemble trust-region policy optimization} (ME-TRPO), which is motivated similarly to this work in that they also propose an algorithm which uses an ensemble of models to mitigate the deleterious effects of model bias. However, the algorithm is quite different. ME-TRPO is a purely model-based policy-gradient approach, and uses the ensemble to avoid overfitting to any one model. In contrast, STEVE interpolates between model-free and model-based estimates, uses a value-estimation approach, and uses the ensemble to explicitly estimate uncertainty.

\citet{kalweit2017uncertainty} train on a mix of real and imagined rollouts, and adjust the ratio over the course of training by tying it to the variance of the Q-function. Similarly to our work, this variance is computed via an ensemble. However, they do not adapt to the uncertainty of individual estimates, only the overall ratio of real to imagined data. Additionally, they do not take into account model bias, or uncertainty in model predictions.

\citet{weber2017imagination} use rollouts generated by the dynamics model as inputs to the policy function, by ``summarizing'' the outputs of the rollouts with a deep neural network. This second network allows the algorithm to implicitly calculate uncertainty over various parts of the rollout and use that information when making its decision. However, I2A has only been evaluated on discrete domains. Additionally, the lack of explicit model use likely tempers the sample-efficiency benefits gained relative to more traditional model-based learning.

\citet{gal2016improving} use a deep neural network in combination with the PILCO algorithm~\cite{deisenroth2011pilco} to do sample-efficient reinforcement learning. They demonstrate good performance on the continuous-control task of cartpole swing-up. They model uncertainty in the learned neural dynamics function using dropout as a Bayesian approximation, and provide evidence that maintaining these uncertainty estimates is very important for model-based reinforcement learning.

\citet{depeweg2016learning} use a Bayesian neural network as the environment model in a policy search setting, learning a policy purely from imagined rollouts. This work also demonstrates that modeling uncertainty is important for model-based reinforcement learning with neural network models, and that uncertainty-aware models can escape many common pitfalls.

\citet{gu2016continuous} propose a continuous variant of Q-learning known as \textit{normalized advantage functions} (NAF), and show that learning using NAF can be accelerated by using a model-based component. They use a variant of Dyna-Q~\cite{sutton1990integrated}, augmenting the experience available to the model-free learner with imaginary on-policy data generated via environment rollouts. They use an iLQG controller and a learned locally-linear model to plan over small, easily-modelled regions of the environment, but find that using more complex neural network models of the environment can yield errors.

\citet{thomas2015policy} define the $\Omega$-return, an alternative to the $\lambda$-return that accounts for the variance of, and correlations between, predicted returns at multiple timesteps. Similarly to STEVE, the target used is an unbiased linear combination of returns with minimum variance. However, the $\Omega$-return is not directly computable in non-tabular state spaces, and does n-step off-policy learning rather than learn a predictive model of the environment. Drawing a theoretical connection between the STEVE algorithm and the $\Omega$-return is an interesting potential direction for future work.
\section{Conclusion}
\label{conclusion}

In this work, we
demonstrated that STEVE, an uncertainty-aware approach for merging model-free and model-based reinforcement learning,
outperforms model-free approaches while reducing sample complexity by an order magnitude on several challenging tasks.
We believe that this is a strong step towards enabling RL for practical, real-world applications.
Since submitting this manuscript for publication, we have further explored the relationship between STEVE and recent work on overestimation bias \cite{pmlr-v80-fujimoto18a}, and found evidence that STEVE may help to reduce this bias.
Other future directions include exploring more complex worldmodels for various tasks, as well as comparing
various techniques for calculating uncertainty and estimating bias.
\subsubsection*{Acknowledgments}

The authors would like to thank the following individuals for their valuable insights and discussion: David Ha, Prajit Ramachandran, Tuomas Haarnoja, Dustin Tran, Matt Johnson, Matt Hoffman, Ishaan Gulrajani, and Sergey Levine. Also, we would like to thank Jascha Sohl-Dickstein, Joseph Antognini, Shane Gu, and Samy Bengio for their feedback during the writing process, and Erwin Coumans for his help on PyBullet enivronments. Finally, we would like to thank our anonymous reviewers for their insightful suggestions.

\bibliography{references}

\begin{thebibliography}{31}
\providecommand{\natexlab}[1]{#1}
\providecommand{\url}[1]{\texttt{#1}}
\expandafter\ifx\csname urlstyle\endcsname\relax
  \providecommand{\doi}[1]{doi: #1}\else
  \providecommand{\doi}{doi: \begingroup \urlstyle{rm}\Url}\fi

\bibitem[Abadi et~al.(2016)Abadi, Barham, Chen, Chen, Davis, Dean, Devin,
  Ghemawat, Irving, Isard, et~al.]{abadi2016tensorflow}
M.~Abadi, P.~Barham, J.~Chen, Z.~Chen, A.~Davis, J.~Dean, M.~Devin,
  S.~Ghemawat, G.~Irving, M.~Isard, et~al.
\newblock Tensorflow: A system for large-scale machine learning.
\newblock In \emph{OSDI}, volume~16, pages 265--283, 2016.

\bibitem[Barth-Maron et~al.(2018)Barth-Maron, Hoffman, Budden, Dabney, Horgan,
  TB, Muldal, Heess, and Lillicrap]{barth-maron2018distributional}
G.~Barth-Maron, M.~W. Hoffman, D.~Budden, W.~Dabney, D.~Horgan, D.~TB,
  A.~Muldal, N.~Heess, and T.~Lillicrap.
\newblock Distributional policy gradients.
\newblock In \emph{International Conference on Learning Representations}, 2018.

\bibitem[Brockman et~al.(2016)Brockman, Cheung, Pettersson, Schneider,
  Schulman, Tang, and Zaremba]{brockman2016openai}
G.~Brockman, V.~Cheung, L.~Pettersson, J.~Schneider, J.~Schulman, J.~Tang, and
  W.~Zaremba.
\newblock Openai gym.
\newblock \emph{arXiv preprint arXiv:1606.01540}, 2016.

\bibitem[Deisenroth and Rasmussen(2011)]{deisenroth2011pilco}
M.~Deisenroth and C.~E. Rasmussen.
\newblock {PILCO}: A model-based and data-efficient approach to policy search.
\newblock In \emph{Proceedings of the 28th International Conference on machine
  learning (ICML-11)}, pages 465--472, 2011.

\bibitem[Depeweg et~al.(2016)Depeweg, Hern{\'a}ndez-Lobato, Doshi-Velez, and
  Udluft]{depeweg2016learning}
S.~Depeweg, J.~M. Hern{\'a}ndez-Lobato, F.~Doshi-Velez, and S.~Udluft.
\newblock Learning and policy search in stochastic dynamical systems with
  {Bayesian} neural networks.
\newblock 2016.

\bibitem[Espeholt et~al.(2018)Espeholt, Soyer, Munos, Simonyan, Mnih, Ward,
  Doron, Firoiu, Harley, Dunning, et~al.]{espeholt2018impala}
L.~Espeholt, H.~Soyer, R.~Munos, K.~Simonyan, V.~Mnih, T.~Ward, Y.~Doron,
  V.~Firoiu, T.~Harley, I.~Dunning, et~al.
\newblock Impala: Scalable distributed deep-rl with importance weighted
  actor-learner architectures.
\newblock In \emph{Proceedings of the International Conference on Machine
  Learning}, 2018.

\bibitem[Feinberg et~al.(2018)Feinberg, Wan, Stoica, Jordan, Gonzalez, and
  Levine]{feinberg2018model}
V.~Feinberg, A.~Wan, I.~Stoica, M.~I. Jordan, J.~E. Gonzalez, and S.~Levine.
\newblock Model-based value estimation for efficient model-free reinforcement
  learning.
\newblock \emph{arXiv preprint arXiv:1803.00101}, 2018.

\bibitem[Fleiss(1993)]{fleiss1993review}
J.~Fleiss.
\newblock Review papers: The statistical basis of meta-analysis.
\newblock \emph{Statistical methods in medical research}, 2\penalty0
  (2):\penalty0 121--145, 1993.

\bibitem[Fujimoto et~al.(2018)Fujimoto, van Hoof, and
  Meger]{pmlr-v80-fujimoto18a}
S.~Fujimoto, H.~van Hoof, and D.~Meger.
\newblock Addressing function approximation error in actor-critic methods.
\newblock In J.~Dy and A.~Krause, editors, \emph{Proceedings of the 35th
  International Conference on Machine Learning}, volume~80 of \emph{Proceedings
  of Machine Learning Research}, pages 1587--1596, Stockholmsmässan, Stockholm
  Sweden, 10--15 Jul 2018. PMLR.
\newblock URL \url{http://proceedings.mlr.press/v80/fujimoto18a.html}.

\bibitem[Gal and Ghahramani(2016)]{gal2016dropout}
Y.~Gal and Z.~Ghahramani.
\newblock Dropout as a {Bayesian} approximation: Representing model uncertainty
  in deep learning.
\newblock In \emph{international conference on machine learning}, pages
  1050--1059, 2016.

\bibitem[Gal et~al.()Gal, McAllister, and Rasmussen]{gal2016improving}
Y.~Gal, R.~McAllister, and C.~E. Rasmussen.
\newblock Improving {PILCO} with {Bayesian} neural network dynamics models.

\bibitem[Gu et~al.(2016)Gu, Lillicrap, Sutskever, and Levine]{gu2016continuous}
S.~Gu, T.~Lillicrap, I.~Sutskever, and S.~Levine.
\newblock Continuous deep {Q}-learning with model-based acceleration.
\newblock In \emph{International Conference on Machine Learning}, pages
  2829--2838, 2016.

\bibitem[Gu et~al.(2017)Gu, Lillicrap, Ghahramani, Turner, and Levine]{gu2016q}
S.~Gu, T.~Lillicrap, Z.~Ghahramani, R.~E. Turner, and S.~Levine.
\newblock Q-prop: Sample-efficient policy gradient with an off-policy critic.
\newblock \emph{International Conference on Learning Representations}, 2017.

\bibitem[Haarnoja et~al.(2018)Haarnoja, Zhou, Abbeel, and
  Levine]{haarnoja2018soft}
T.~Haarnoja, A.~Zhou, P.~Abbeel, and S.~Levine.
\newblock Soft actor-critic: Off-policy maximum entropy deep reinforcement
  learning with a stochastic actor, 2018.

\bibitem[Heess et~al.(2015)Heess, Wayne, Silver, Lillicrap, Erez, and
  Tassa]{heess2015learning}
N.~Heess, G.~Wayne, D.~Silver, T.~Lillicrap, T.~Erez, and Y.~Tassa.
\newblock Learning continuous control policies by stochastic value gradients.
\newblock In \emph{Advances in Neural Information Processing Systems}, pages
  2944--2952, 2015.

\bibitem[Horgan et~al.(2018)Horgan, Quan, Budden, Barth-Maron, Hessel, van
  Hasselt, and Silver]{horgan2018distributed}
D.~Horgan, J.~Quan, D.~Budden, G.~Barth-Maron, M.~Hessel, H.~van Hasselt, and
  D.~Silver.
\newblock Distributed prioritized experience replay.
\newblock In \emph{International Conference on Learning Representations}, 2018.

\bibitem[Kalweit and Boedecker(2017)]{kalweit2017uncertainty}
G.~Kalweit and J.~Boedecker.
\newblock Uncertainty-driven imagination for continuous deep reinforcement
  learning.
\newblock In \emph{Conference on Robot Learning}, pages 195--206, 2017.

\bibitem[Kingma and Ba(2015)]{kingma2014adam}
D.~P. Kingma and J.~Ba.
\newblock Adam: A method for stochastic optimization.
\newblock \emph{International Conference on Learning Representations}, 2015.

\bibitem[Klimov and Schulman()]{roboschool}
O.~Klimov and J.~Schulman.
\newblock Roboschool.
\newblock \url{https://github.com/openai/roboschool}.

\bibitem[Kurutach et~al.(2018)Kurutach, Clavera, Duan, Tamar, and
  Abbeel]{kurutach2018modelensemble}
T.~Kurutach, I.~Clavera, Y.~Duan, A.~Tamar, and P.~Abbeel.
\newblock Model-ensemble trust-region policy optimization.
\newblock In \emph{International Conference on Learning Representations}, 2018.

\bibitem[Lillicrap et~al.(2016)Lillicrap, Hunt, Pritzel, Heess, Erez, Tassa,
  Silver, and Wierstra]{lillicrap2015continuous}
T.~P. Lillicrap, J.~J. Hunt, A.~Pritzel, N.~Heess, T.~Erez, Y.~Tassa,
  D.~Silver, and D.~Wierstra.
\newblock Continuous control with deep reinforcement learning.
\newblock \emph{International Conference on Learning Representations}, 2016.

\bibitem[MacKay(1992)]{mackay1992practical}
D.~J. MacKay.
\newblock A practical {Bayesian} framework for backpropagation networks.
\newblock \emph{Neural computation}, 4\penalty0 (3):\penalty0 448--472, 1992.

\bibitem[Mnih et~al.(2013)Mnih, Kavukcuoglu, Silver, Graves, Antonoglou,
  Wierstra, and Riedmiller]{mnih2013playing}
V.~Mnih, K.~Kavukcuoglu, D.~Silver, A.~Graves, I.~Antonoglou, D.~Wierstra, and
  M.~Riedmiller.
\newblock Playing atari with deep reinforcement learning.
\newblock In \emph{NIPS Deep Learning Workshop}. 2013.

\bibitem[Munos et~al.(2016)Munos, Stepleton, Harutyunyan, and
  Bellemare]{munos2016safe}
R.~Munos, T.~Stepleton, A.~Harutyunyan, and M.~Bellemare.
\newblock Safe and efficient off-policy reinforcement learning.
\newblock In \emph{Advances in Neural Information Processing Systems}, pages
  1054--1062, 2016.

\bibitem[Osband et~al.(2016)Osband, Blundell, Pritzel, and
  Van~Roy]{osband2016deep}
I.~Osband, C.~Blundell, A.~Pritzel, and B.~Van~Roy.
\newblock Deep exploration via bootstrapped {DQN}.
\newblock In \emph{Advances in neural information processing systems}, pages
  4026--4034, 2016.

\bibitem[Schulman et~al.(2017)Schulman, Wolski, Dhariwal, Radford, and
  Klimov]{schulman2017proximal}
J.~Schulman, F.~Wolski, P.~Dhariwal, A.~Radford, and O.~Klimov.
\newblock Proximal policy optimization algorithms.
\newblock \emph{arXiv preprint arXiv:1707.06347}, 2017.

\bibitem[Silver et~al.(2016)Silver, Huang, Maddison, Guez, Sifre, Van
  Den~Driessche, Schrittwieser, Antonoglou, Panneershelvam, Lanctot,
  et~al.]{silver2016mastering}
D.~Silver, A.~Huang, C.~J. Maddison, A.~Guez, L.~Sifre, G.~Van Den~Driessche,
  J.~Schrittwieser, I.~Antonoglou, V.~Panneershelvam, M.~Lanctot, et~al.
\newblock Mastering the game of go with deep neural networks and tree search.
\newblock \emph{nature}, 529\penalty0 (7587):\penalty0 484--489, 2016.

\bibitem[Sutton(1990)]{sutton1990integrated}
R.~S. Sutton.
\newblock Integrated architectures for learning, planning, and reacting based
  on approximating dynamic programming.
\newblock In \emph{Machine Learning Proceedings 1990}, pages 216--224.
  Elsevier, 1990.

\bibitem[Sutton and Barto(1998)]{sutton1998reinforcement}
R.~S. Sutton and A.~G. Barto.
\newblock \emph{Reinforcement Learning: An Introduction}, volume~1.
\newblock MIT Press Cambridge, 1998.

\bibitem[Thomas et~al.(2015)Thomas, Niekum, Theocharous, and
  Konidaris]{thomas2015policy}
P.~S. Thomas, S.~Niekum, G.~Theocharous, and G.~Konidaris.
\newblock Policy evaluation using the {$\Omega$}-return.
\newblock In \emph{Advances in Neural Information Processing Systems}, pages
  334--342, 2015.

\bibitem[Weber et~al.(2017)Weber, Racani{\`e}re, Reichert, Buesing, Guez,
  Rezende, Badia, Vinyals, Heess, Li, et~al.]{weber2017imagination}
T.~Weber, S.~Racani{\`e}re, D.~P. Reichert, L.~Buesing, A.~Guez, D.~J. Rezende,
  A.~P. Badia, O.~Vinyals, N.~Heess, Y.~Li, et~al.
\newblock Imagination-augmented agents for deep reinforcement learning.
\newblock \emph{31st Conference on Neural Information Processing Systems},
  2017.

\end{thebibliography}

\clearpage
\appendix
\section{Toy Problem: A Tabular FSM with Model Noise}
\label{toy}

To demonstrate the benefits of Bayesian model-based value expansion, we evaluated it on a toy problem. We used a finite state environment with states $\{s_0, \ldots, s_{100} \}$, and a single action $A$ available at every state which always moves from state $s_t$ to $s_{t+1}$, starting at $s_0$ and terminating at $s_{100}$. The reward for every action
is -1, except when moving from $s_{99}$ to $s_{100}$, which is +100.
Since this environment is so simple, there is only one possible policy $\pi$, which is deterministic. It is possible to compute
the true action-value function in closed form, which is $Q^\pi(s_i, A) = i$.

We estimate the value of each state using tabular TD-learning. We maintain a tabular function $\hat{Q}^\pi(s_i, A)$, which is just a lookup table matching each state to its estimated value. We initialize all values to random integers between 0 and 99, except for the terminal state $s_{100}$, which we initialize to 0 (and keep fixed at 0 at all times). We update using the standard undiscounted one-step TD update, $\hat{Q}^\pi(s_i, A) = r + \hat{Q}^\pi(s_{i+1}, A)$. For each update, we sampled a nonterminal state and its corresponding transition $(s_i, r, s_{i+1})$ at random.
For experiments with an ensemble of Q-functions, we repeat this update once for each member of the ensemble at each timestep.

The transition and reward function for the oracle dynamics model behaved exactly the same as the true environment.
In the ``noisy'' dynamics model, noise was added in the following way: 10\% of the time, rather than correctly moving from $s_t$
to $s_{t+1}$, the model transitions to a random state. (Other techniques for adding noise gave qualitatively similar results.)

On the y-axis of Figure \ref{fig:toy}, we plot the mean squared error between the predicted values and the true values of each state: $\frac{1}{100} \sum_{i=0}^{99} (\hat{Q}^\pi(s_i, A) - Q^\pi(s_i, A))^2$.

For both the STEVE and MVE experiments, we use an ensemble of size 8 for both the model and the Q-function. To compute the MVE target, we average across all ensembled rollouts and predictions.

\section{The TD-k Trick}
\label{appendix:tdk}

The TD-k trick, proposed by \citet{feinberg2018model}, involves training the Q-function using every intermediate state of the rollout:

\begin{align*}
    s'_{-1} &= s \\
    \mathcal{L}_\theta &= \E_{(s, a, r, s')}\left[ \frac{1}{H} \sum_{i=-1}^{H-1} (\hat{Q}_\theta^\pi(s'_i, a_i) - \mathcal{T}_H^{\MVE}(r_i,s'_{i+1}))^2 \right],
\end{align*}

where $s'_i, r_i, a_i$ are defined as in Equation \ref{mve_rollout_formula}.

To summarize \citet{feinberg2018model}, the TD-k trick is helpful because the off-policy states collected by the replay buffer may have little overlap with the states encountered during on-policy model rollouts. Without the TD-k trick, the Q-function approximator is trained to minimize error only on states collected from the replay buffer, so it is likely to have high error on states not present in the replay buffer. This would imply that the Q-function has high error on states produced by model rollouts, and that this error may in fact continue to increase the more steps of on-policy rollout we take. By invoking the TD-k trick, and training the Q-function on intermediate steps of the rollout, \citet{feinberg2018model} show that we can decrease the Q-function bias on frames encountered during model-based rollouts, leading to better targets and improved performance.

The TD-k trick is orthogonal to STEVE. STEVE tends to ignore estimates produced by states with poorly-learned Q-values, so it is not hurt nearly as much as MVE by the distribution mismatch problem. However, better Q-values will certainly provide more information with which to compute STEVE's target, so in that regard the TD-k trick seems beneficial. An obvious question is whether these two approaches are complimentary.
STEVE+TD-k is beyond the scope of this work, and we did not give it a rigorous treatment; however, initial experiments were not promising. In future work, we hope to explore the connection between these two approaches more deeply.

\section{Implementation Details}
\label{implement}

All models were feedforward neural networks with ReLU nonlinearities. The policy network, reward model, and termination model each had 4 layers of size 128, while the transition model had 8 layers of size 512. All environments were reset after 1000 timesteps. Parameters were trained with the Adam optimizer \citet{kingma2014adam} with a learning rate of 3e-4.

Policies were trained using minibatches of size 512 sampled uniformly at random from a replay buffer of size 1e6. The first 1e5 frames were sampled via random interaction with the environment; after that, 4 policy updates were performed for every frame sampled from the environment. (In Section \ref{experiment:wallclock}, the policy updates and frames were instead de-synced.) Policy checkpoints were saved every 500 updates; these checkpoints were also frozen and used as $\theta^-$. For model-based algorithms, the most recent checkpoint of the model was loaded every 500 updates as well.

Each policy training had 8 agents interacting with the environment to send frames back to the replay buffer. These agents typically took the greedy action predicted by the policy, but with probability $\epsilon=0.05$, instead took an action sampled from a normal distribution surrounding the pre-tanh logit predicted by the policy. In addition, each policy had two greedy agents interacting with the environment for evaluation.

Dynamics models were trained using minibatches of size 1024 sampled uniformly at random from a replay buffer of size 1e6. The first 1e5 frames were sampled via random interaction with the environment; the dynamics model was then pre-trained for 1e5 updates. After that, 4 model updates were performed for every frame sampled from the environment. (In Section \ref{experiment:wallclock}, the model updates and frames were instead de-synced.) Model checkpoints were saved every 500 updates.

All ensembles were of size 4. During training, each ensemble member was trained on an independently-sampled minibatch; all minibatches were drawn from the same buffer. Additionally, $M,N,L = 4$ for all experiments.

\end{document}